\documentclass[letterpaper]{article} 
\usepackage[misc]{ifsym}
\usepackage{aaai2026}  
\usepackage{times} 
 \usepackage{helvet}  
 \usepackage{courier}  
 \usepackage[hyphens]{url}  
 \usepackage{graphicx}    \usepackage{natbib}  \usepackage{caption}       \usepackage{algorithm}
\usepackage{microtype}
\usepackage{xcolor}
\usepackage{graphicx}
\usepackage{subcaption}
\usepackage{booktabs} \usepackage{amsmath}
\usepackage{amssymb}
\usepackage{multirow}
\usepackage{enumitem}
\usepackage{algpseudocode}
\usepackage{array}
\usepackage{amsmath,amsfonts,bm}

\usepackage{amsthm}

\newtheorem{Theorem}{Theorem}

\newtheorem{Definition}{Definition}

\def\eqref#1{equation~\ref{#1}}
\def\1{\bm{1}}

\def\vh{{\bm{h}}}

\def\vp{{\bm{p}}}

\def\vz{{\bm{z}}}

\def\mA{{\bm{A}}}

\def\mM{{\bm{M}}}

\def\mX{{\bm{X}}}

\DeclareMathAlphabet{\mathsfit}{\encodingdefault}{\sfdefault}{m}{sl}
\SetMathAlphabet{\mathsfit}{bold}{\encodingdefault}{\sfdefault}{bx}{n}

\def\gE{{\mathcal{E}}}

\def\gT{{\mathcal{T}}}

\def\gV{{\mathcal{V}}}

\def\emM{{M}}

\DeclareMathOperator*{\argmax}{arg\,max}
\DeclareMathOperator*{\argmin}{arg\,min}

\newcommand{\bamo}{BA-2motifs}
\newcommand{\prot}{PROTEINS}
\newcommand{\benz}{Benzene}
\newcommand{\mutag}{MUTAG}
\newcommand{\fluo}{Fluoride-Carbonyl}
\newcommand{\alk}{Alkane-Carbonyl}
\newcommand{\dd}{D\&D}
\newcommand{\hiv}{HIV}
\newcommand{\RB}{REDDIT-BINARY}
\newcommand{\ours}{\textsc{EPA}}

\usepackage{newfloat}
\usepackage{listings}
\DeclareCaptionStyle{ruled}{labelfont=normalfont,labelsep=colon,strut=off} 
\floatstyle{ruled}
\newfloat{listing}{tb}{lst}{}
\floatname{listing}{Listing}
\title{Explanation-Preserving Augmentation for Semi-Supervised Graph Representation Learning}
\author{
    Zhuomin Chen\textsuperscript{\rm 1},
    Jingchao Ni\textsuperscript{\rm 2}, 
    Hojat Allah Salehi\textsuperscript{\rm 1},
    Xu Zheng\textsuperscript{\rm 1},
    Esteban Schafir\textsuperscript{\rm 1},
    Farhad Shirani \thanks{Farhad Shirani was affiliated with Florida International University during the preparation of this work.},
    Dongsheng Luo\textsuperscript{\rm 1 \Letter}
}
\affiliations{
        \textsuperscript{\rm 1} Knight Foundation School of Computing and Information
Sciences, Florida International University\\
            
                        \textsuperscript{\rm 2} Department of
Computer Science, University of Houston \\

        zchen051@fiu.edu, 
    jni7@uh.edu,
    hsalehi@fiu.edu, 
    xzhen019@fiu.edu, 
    escha032@fiu.edu,
    luodongsheng01@gmail.com
}

\usepackage{bibentry}

\begin{document}

\maketitle

\begin{abstract}
Self-supervised graph representation learning (GRL) typically generates paired graph augmentations from each graph to infer similar representations for augmentations of the same graph, but distinguishable representations for different graphs. While effective augmentation requires both semantics-preservation and data-perturbation, most existing GRL methods focus solely on data-perturbation, leading to suboptimal solutions.  To fill the gap, in this paper, we propose a novel method, Explanation-Preserving Augmentation (\ours), which leverages graph explanation for semantics-preservation. 
{\ours} first uses a small number of labels to train a graph explainer, which infers the 
subgraphs that explain the graph's label. Then these explanations are used for generating semantics-preserving augmentations for boosting self-supervised GRL. Thus, the entire process, namely \ours-GRL, is semi-supervised. We demonstrate theoretically, using an analytical example, and through extensive experiments on 
a variety of benchmark datasets, that \ours-GRL outperforms the state-of-the-art (SOTA) GRL methods that use semantics-agnostic augmentations. The code is available at~\url{https://github.com/realMoana/EPA-GRL}.
\end{abstract}

\section{Introduction}\label{sec.introduction}

Inspired by recent progress in self-supervised representation learning in vision and language domains \cite{chen2020simple}, contrastive learning has emerged as a predominant technique for graph representation learning (GRL). Typically, two augmentations are generated for each graph with the objective of learning similar representations for augmentations of the same graph but discriminative representations for augmentations of different graphs. The success of self-supervised GRL is grounded in an effective augmentation strategy. Analogous to image data augmentation \cite{he2020momentum}, an ideal pair of graph augmentations should concurrently be able to (1) inherit the semantics -- which may be represented by signature subgraphs pertinent to the classification -- of their original graph; and (2) present sufficient variance from each other~\cite{yin2022autogcl}. 

However, most existing works only focus on structural perturbations that add variance to the augmented graphs but largely neglect the need for preserving semantics. For example, in graph contrastive learning (GraphCL) \cite{graphcl} and JOAO \cite{you2021graph}, an augmented graph is typically generated by perturbing its original graph through random node/edge dropping, feature masking, and subgraph extraction. The randomness in these perturbations inevitably induces substantial alterations to some important (sub-)graph structures or features that may result in a considerable loss of semantics, which in turn, may lead to significant performance drops in downstream tasks.

\begin{figure}[t]
    \centering
    \includegraphics[width=0.9\linewidth]{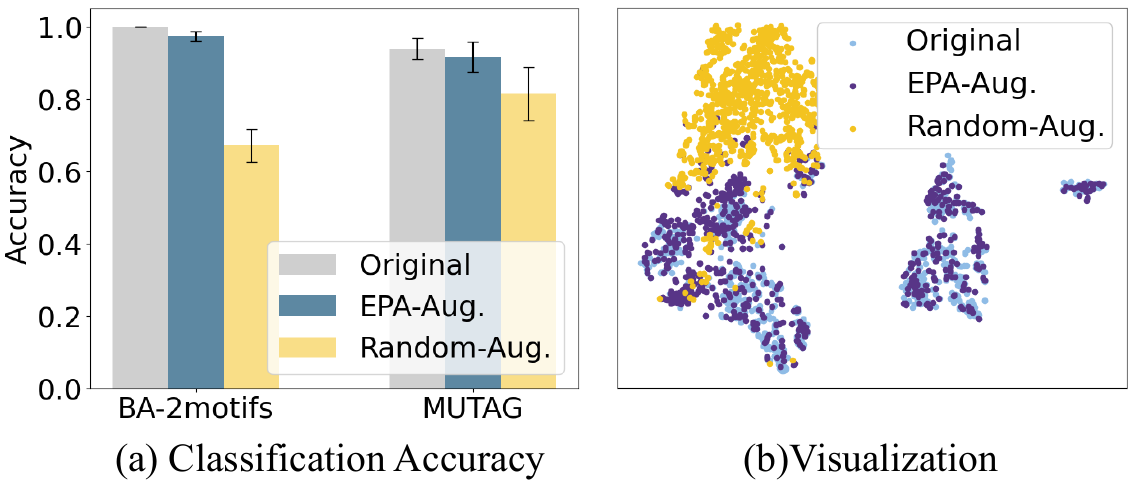}
    \vspace{-0.1cm}
    \caption{Semantics-preserving ability of different augmentations.}
    \label{fig:augmentation_comparison}
\end{figure}

Fig.~\ref{fig:augmentation_comparison} illustrates an experiment that evaluates the semantics-preserving ability of different augmentation techniques. A Graph Neural Network (GNN) classifier \cite{kipf2016semi} is first trained on the training partition of a benchmark dataset ({\em i.e.}, BA-2motifs~\cite{luo2020parameterized} or MUTAG~\cite{debnath1991structure}),
and its accuracy is then evaluated on the original graphs in the test set and the augmented graphs derived from them, respectively. Two augmentation methods are included. ``Random-Aug'' represents randomly dropping nodes from the original graph in a {\em semantics-agnostic} manner. ``\ours-Aug'' is an explanation-preserving augmentation (\ours) that operates in a semantics-preserving manner (which will be introduced later). From Fig. \ref{fig:augmentation_comparison}(a), the fully trained GNN can accurately classify the graphs in the original test set.
However, it has a sharp drop in accuracy on the ``Random-Aug'' graphs, suggesting a significant loss of semantics ({\em i.e.}, class-related subgraphs).

In contrast, the accuracy on ``\ours-Aug'' graphs is close to that of the original graphs, implying effective semantics-preservation. Fig.\ref{fig:augmentation_comparison}(b) shows the graph embeddings in BA-2motifs. A large distribution shift from the original graphs to the ``Random-Aug'' graphs can be observed. In addition, ``\ours-Aug'' well preserves the distribution of the original embeddings while presenting sufficient variance. Thus, semantics-preserving augmentations are more suitable than semantics-agnostic ones for training a generalizable GNN using GRL.

Prior attempts to address semantics preservation either rely on domain-specific knowledge \cite{sun2021mocl}, which limits generalizability, or employ parameter perturbations \cite{xia2022simgrace} that may implicitly alter graph structures. Some methods use unsupervised approaches to preserve structural patterns \cite{shi2023engage}, but without label guidance, these patterns may be irrelevant to class discrimination, limiting their effectiveness for downstream tasks.

In this paper, we aim to develop semantics-preserving augmentation techniques -- using models trained by leveraging only a few labeled input samples -- to enhance GRL.
Inspired by recent research on explainable AI (XAI) for graphs \cite{ying2019gnnexplainer,luo2020parameterized,yuan2022explainability}, we propose a novel approach, Explanation-Preserving Augmentation enhanced GRL ({\ours-GRL}), 
which leverages graph explanation techniques for generating augmented graphs that can bridge the gap between semantics-preservation and data-perturbation. Methods for explaining GNNs usually learn a parametric {\em graph explainer} that can identify a sub-structure ({\em e.g.}, benzene ring) in the original graph ({\em e.g.}, molecule) that distinguishes the graph from the graphs of other classes \cite{luo2020parameterized}. In other words, the explainer infers semantics represented by subgraph patterns. In light of this, \ours-GRL is designed as a two-stage approach. At the pre-training stage, it learns a graph explainer using a handful of class labels. At the representation learning stage, for each input graph, \ours-GRL uses the explainer to extract a semantic subgraph and perturbs the rest of the original graphs ({\em i.e.}, marginal subgraph). The semantics subgraph and the perturbed marginal subgraph are combined to form an augmented graph, which is fed to a contrastive learning framework for representation learning. In this way, \ours-GRL uses a few labeled graphs and relatively more unlabeled graphs, establishing a novel label-efficient semi-supervised GRL framework. 
In summary, the main contributions are as follows.
\begin{itemize}[leftmargin=*]
\setlength{\itemsep}{-0.5pt}
    \item We identify a key limitation of the existing GRL augmentation methods as per the criteria -- {\em semantics-preservation} and {\em data-perturbation} -- of data augmentation.
    \item This work is the first to explore the potential of a few class labels in semantics-preservation for GRL. We propose a new semi-supervised GRL framework with a novel augmentation method \ours, which introduces XAI to GRL for semantics-preserving perturbation.
    \item We show theoretically, via an analytical example, that by operating on the output embeddings of a GRL model, the accuracy gap of an empirical risk minimizer under semantics-preserving and semantics-agnostic augmentations can be arbitrarily large.
    \item We conduct experiments on 6 benchmark datasets with extensive comparison of augmentation methods and contrastive learning frameworks. The results validate the effectiveness of the proposed \ours-GRL method, especially when labeled data are limited. 
        
\end{itemize}

\begin{figure*}[ht]
    \centering
    \includegraphics[width=0.9\textwidth]{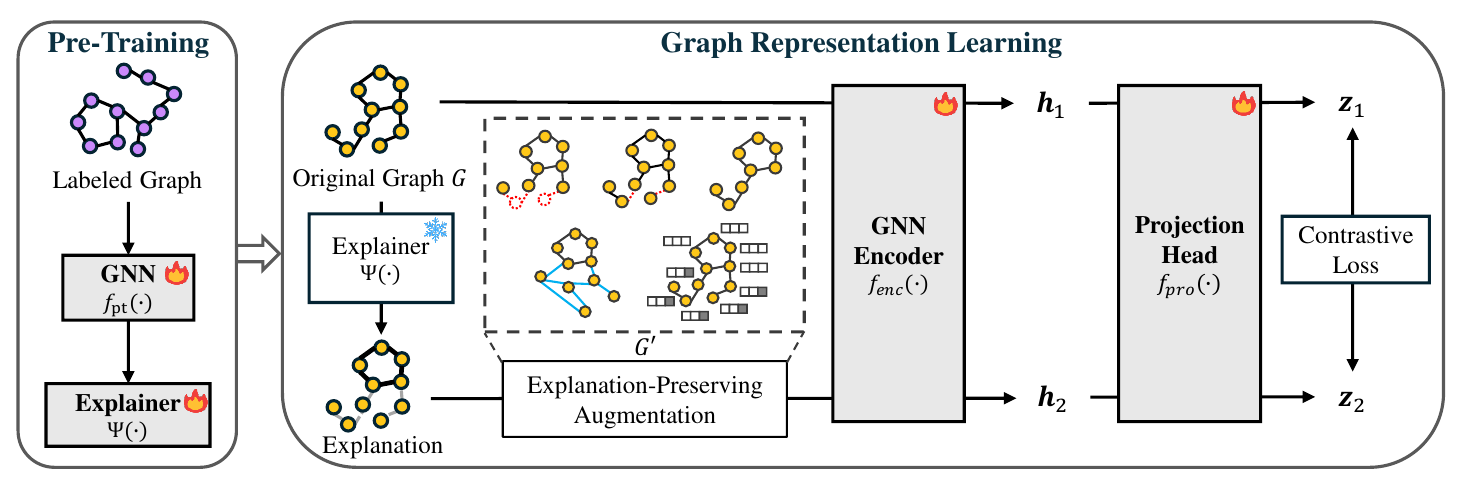}
    \caption{The Architecture of the proposed {\ours-GRL} method. We first pretrain a GNN model $f_{\text{pt}}(\cdot)$ and its explainer $\Psi(\cdot)$ with a small number of labeled training samples. Then in the GRL step, we use the frozen explainer $\Psi(\cdot)$ to produce augmented graphs to train a GNN encoder $f_\text{enc}(\cdot)$ and a projection head $f_\text{pro}(\cdot)$ with a contrastive loss. The output of the GNN encoder $f_\text{enc}(\cdot)$ will be used as graph representations.} 
    \label{fig:epamodel}
\end{figure*}

\section{Related Work}\label{sec.relatedwork}

\noindent \textbf{Graph Contrastive Learning.}
Contrastive learning on graphs includes node-level \cite{zhu2020deep, zhu2021graph} and graph-level tasks \cite{sun2019infograph, graphcl, you2021graph, suresh2021adversarial, yin2022autogcl, xia2022simgrace}. This study focuses on graph-level tasks.  GraphCL \cite{graphcl} performs graph contrastive learning using four different data augmentation methods: node dropping, edge dropping, subgraph sampling, and attribute masking; JOAO \cite{you2021graph} propose a unified bi-level optimization framework that automatically selects the suitable data augmentation method from GraphCL; AD-GCL \cite{suresh2021adversarial} utilizes adversarial augmentation methods to prevent GNNs from capturing redundant information from the original graph during training; AutoGCL \cite{yin2022autogcl} uses learnable graph view generators guided by an automated augmentation strategy to introduce appropriate augmentation variances during contrastive learning; SimGRACE \cite{xia2022simgrace} uses different graph encoders as generators of contrastive graphs and compares the semantic similarity between graphs obtained from the perturbed encoders for contrastive learning.
DRGCL \cite{ji2024rethinking} generates augmented graphs by randomly retaining certain representation dimensions and refines them using learnable, dimension-specific weights. CI-GCL \cite{tan2024community} proposes a community-invariant contrastive learning framework
by leveraging community information to construct positive pairs. \\
\textbf{Explainable Graph Neural Networks.}
Explainability in GNNs has gained significant attention, with various methods proposed to enhance transparency in graph-based tasks \cite{ying2019gnnexplainer, luo2020parameterized, yuan2020xgnn, yuan2021explainability, wang2022gnninterpreter, xie2022task, ma2022clear, miao2023interpretable, fang2023cooperative,  zheng2024robustfidelityevaluatingexplainability, chen2024generating}. 
These approaches improve understanding of GNN decisions at both instance \cite{ying2019gnnexplainer, luo2020parameterized, wang2021towards, zhang2023mixupexplainer, chen2024generating} and model levels \cite{yuan2020xgnn, wang2022gnninterpreter, shin2024pageprototypebasedmodellevelexplanations}. Early methods like Saliency Maps \cite{baldassarre2019explainability} and Grad-CAM \cite{pope2019explainability} relied on gradients, while recent advances introduced perturbation-based methods \cite{luo2020parameterized, wang2021towards}, surrogate models \cite{vu2020pgm, duval2021graphsvx}, and generation-based approaches \cite{yuan2020xgnn, shan2021reinforcement, wang2022gnninterpreter}. Perturbation-based methods, such as GNNExplainer \cite{ying2019gnnexplainer}, PGExplainer \cite{luo2020parameterized}, Refine \cite{wang2021towards}, MixupExplainer \cite{zhang2023mixupexplainer}, and ProxyExplainer \cite{chen2024generating}, generate explanations by perturbing graph features or structures to identify the most important components influencing predictions. 
Surrogate models \cite{vu2020pgm}  approximate the original GNN with simpler models to explain local predictions, while generation-based methods \cite{yuan2020xgnn, wang2022gnninterpreter}  leverage generative models to provide both instance-level and model-level explanations.

Some recent works have attempted to use explanation to improve learning performance~\cite{shi2023engage,wang2021molecular}. For instance, ENGAGE \cite{shi2023engage} proposes a Smoothed Activation Map to identify important nodes based on representation distributions and uses this to guide graph augmentation in contrastive learning. However, without access to semantic supervision, such unsupervised explanations can only capture structural patterns (e.g., nodes with similar local neighborhoods or high connectivity) that may not align with class-discriminative features. In contrast, our work leverages supervised explanation with limited labels to identify and preserve 
truly class-relevant semantic structures during augmentation.

\section{Notations and Problem Formulation}
A graph \( G \) is defined by:
i) a node set \( \mathcal{V} = \{v_1, v_2, \dots, v_n\} \), where $n$ is the number of nodes; 
ii) an edge set \( \mathcal{E} \subseteq \mathcal{V} \times \mathcal{V} \);  
iii) a feature matrix \( \mX \in \mathbb{R}^{n \times d} \), where the \( i \)-th row \( \mX_i \) is the $d$-dimensional feature of node \( v_i \); and iv) an adjacency matrix \( \mA \in \{0,1\}^{n \times n} \), where \( A_{i,j} = 1\) if \((v_i, v_j) \in \mathcal{E}\). 
Additionally, each graph is associated with a label \( Y \in \mathcal{Y} \), where \( \mathcal{Y} \) is a finite set.

Formally, we assume a pair of training sets, a (small) labeled set $\mathcal{T}_{\ell}= \{(G_i,Y_i)|i\in [M]\}$, and a (large) unlabeled set $\mathcal{T}_{u}= \{G_i|i\in [N]\}$, where $M\ll N$. 
Our objective is to leverage the small number of labeled graphs in $\mathcal{T}_{\ell}$ and relatively more unlabeled graphs in $\mathcal{T}_{u}$ to perform semantics-preserving representation learning. Similar to the supervised contrastive learning methods 
\cite{khosla2020supervised,ji2023supervised}, labels from a classification task are leveraged for model training. However, {\em our problem is substantially different from the existing works by only using a few labels, leading to a constrained semi-supervised GRL problem.}

\section{The Proposed Method}
In this section, we introduce the proposed \ours-GRL method. 
Fig. \ref{fig:epamodel} is an overview. 
First, \ours-GRL pre-trains an 
explainer using the labeled graphs in $\mathcal{T}_{\ell}$, where the explainer $\Psi(\cdot):G\mapsto G^{\text{(exp)}}$ takes a graph $G$ as input and outputs an explanation subgraph $G^{\text{(exp)}}$ which is class-specific. Then, a semantic-preserving stochastic mapping $P_{G'|G}$ transforms the original graph $G$ to an augmentation $G'$ such that $G^{\text{(exp)}}\subseteq G'$, 
where the mapping can perform random perturbations on the non-explanatory ({\em i.e.}, marginal) part of the input $G$, {\em i.e.}, $G\setminus G^{\text{(exp)}}$; Finally, the augmented graph $G'$ is fed to an encoder $f_\text{enc}:G' \mapsto \vh$, where $\vh \in \mathbb{R}^{d_e}$ is the $d_e$-dimensional embedding of $G'$. 
$f_\text{enc}$ is trained via contrastive learning on both augmentations of the unlabeled training set $\mathcal{T}_u$ and the labeled training set $\mathcal{T}_\ell$.

\subsection{Pre-Training GNN Explainer}
In this step, the goal is to train a GNN explainer to identify the most responsible subgraph for its predictions. Extracting such key substructures enables preservation of core semantics of the input graph when perturbing the rest of the graph. 
We begin by training a GNN $f_\text{pt}(\cdot)$ using the labeled training set. The GNN learns to capture both structural and feature-based information necessary to distinguish different graph classes. The GNN is trained by minimizing the cross-entropy loss between the predicted label $f_\text{pt}(G)$ and the ground-truth label $y$. Formally, the optimization problem is:
\begin{equation} 
\label{eq:ce}
\argmin_{f_\text{pt}} \left( {\sum}_{(G,y) \in \gT_\ell} -y \log(f_\text{pt}(G)) \right). 
\end{equation}

Next, we propose to use a GNN explainer to extract subgraphs that retain the semantics necessary for classification. Augmentations can then be achieved by making controlled perturbations to rest, marginal parts of the graph. 
Formally, given a graph $G = (\mA, \mX)$, the explainer generates an explanation subgraph $G^{(\text{exp})} = (\mA \odot \mM, \mX)$, where $\mM \in \{0,1\}^{|\gV| \times |\gV|}$ is a binary mask and each entry $\emM_{ij} = 1$ indicates edge $(i,j)$ is retained in the subgraph $G^{(\text{exp})}$. 
To improve efficiency, we use a generative explainer, $\Psi(\cdot)$, which employs a parametric neural network to learn the mask $\mM$ based on node embeddings~\cite{luo2020parameterized,luo2024towards}. The explainer $\Psi(\cdot)$ is trained on the labeled dataset $\gT_\ell$ and can be directly applied to the unlabeled graphs in $\gT_u$ to generate new explanation subgraphs.

The objective of the GNN explainer $\Psi(\cdot)$ is to find a subgraph, denoted by $\Psi(G)$, that balances semantic preservation and compactness. This is achieved by following the Graph Information Bottleneck (GIB) principle, which has been widely used in GNN explanation methods~\cite{yu2020graph,xu2021infogcl,suresh2021adversarial,yin2022autogcl}. 
The GIB principle states that an optimal subgraph should retain sufficient information for a prediction task while being as compact as possible, to avoid overfitting or including irrelevant parts of the graph. The learning objective for the explainer is defined as: 
\begin{equation} 
\label{eq:explain} 
\argmin_{\Psi} \left( {\sum}_{(G,y) \in \gT_\ell} \text{CE}(Y;f_\text{pt}(\Psi(G))) + \lambda |\Psi(G)| \right ), 
\end{equation}
where $\text{CE}(Y; f_\text{pt}(\Psi(G)))$ is the cross-entropy loss between the label and the prediction on the subgraph $\Psi(G)$, and $|\Psi(G)|$ is the size of the subgraph, which can be measured by the number of edges or the sum of edge weights. The hyper-parameter $\lambda$ controls the trade-off between the terms for information preservation and structural compactness.
\subsection{Explanation-Preserving Augmentation Enhanced Graph Representation Learning}
Data augmentation is essential
in contrastive learning, which generates multiple augmented views of the same graph for learning invariant representations. The success of graph data augmentation attributes to its ability to preserve the core semantics of the graph while introducing variances that facilitate robust representation learning. However, a major limitation of the existing methods is their unconstrained perturbation techniques that may arbitrarily modify the graph structure or node features, and inadvertently disconnect important substructures, leading to a significant loss of semantic information.

\noindent \textbf{Explanation-Preserving Augmentation.} To address this limitation, we propose an EPA strategy that explicitly retains the essential part of the graph regarding its class. Specifically, we use the pre-trained graph explainer $\Psi(\cdot)$ to extract an explanation subgraph $G^{(\text{exp})} = \Psi(G)$, which contains the most relevant substructures to the graph's semantics, and $G^{(\text{exp})}$ will be kept intact. The remaining part of the graph, denoted as the {\em marginal subgraph} $\Delta G = G \setminus G^{(\text{exp})}$, is perturbed to introduce necessary variance for contrastive learning. Next, we introduce EPA-based methods for graph-structured data and discuss their intuitive priors. Detailed algorithms are provided in Appendix B.
\begin{itemize}[leftmargin=*]
    \item \textbf{Node Dropping}     randomly removes a subset of nodes and their edges from the     marginal subgraph $\Delta G$. The assumption is that removing irrelevant nodes has a small impact on the core semantics of the graph. Each node’s dropping probability follows an i.i.d. uniform distribution.
    \item \textbf{Edge Dropping}     modifies the connectivity in $G$ by randomly dropping edges in the     marginal subgraph $\Delta G$. It assumes the semantic meaning of the graph is robust to the changes of inessential edges. Each edge dropping follows an i.i.d. uniform distribution.
    \item \textbf{Attribute Masking} hides a subset of node or edge attributes in the     marginal subgraph $\Delta G$. The assumption is related to node dropping --- missing unimportant node attributes has minor impacts on the recovery of essential semantics by     the explanation subgraph.
    \item \textbf{Subgraph} further samples a subgraph from the     marginal subgraph $\Delta G$ based on the assumption that sampling a connected subgraph in $\Delta G$ varies the graph structure but     does not disrupt the overall semantic integrity.
    \item \textbf{Mixup} randomly selects another graph $\tilde{G}$ from the batch, and combines its marginal subgraph $\Delta \tilde{G}$ with the explanation subgraph $G^{(\text{exp})}$ of the original graph $G$.
\end{itemize}

\noindent \textbf{Graph Representation Learning}. After generating the augmentations, we employ a contrastive learning framework to learn graph representations. Our \ours\ approach is versatile and can be integrated into various graph contrastive learning methods. Here, we demonstrate its flexible applicability with two popular techniques: GraphCL~\cite{graphcl} and SimSiam~\cite{chen2021exploring}.

For each graph $G$ 
we generate an augmented view $G'$ with our \ours\ method. 
These graphs are then passed to a graph encoder $f_\text{enc}(\cdot)$, which can be any suitable GNN architecture. The encoder produces graph-level representations $\vh_1 = f_\text{enc}(G)$ and $\vh_2 = f_\text{enc}(G')$ for the two views. Following the approach in~\cite{chen2020simple}, a projection head, implemented as an MLP, is adopted to obtain new representations for defining the self-supervised learning loss. Specifically, we compute $\vz_1 = f_\text{pro}(\vh_1)$ and $\vz_2 = f_\text{pro}(\vh_2)$. In line with the previous works~\cite{chen2020simple,chen2021exploring}, $\vz_1$ and $\vz_2$ are used exclusively for model training. For downstream tasks, we discard the projection head and utilize $\vh_1$ and $\vh_2$ as the graph representations.

\noindent \textbf{GraphCL} \cite{graphcl} is a contrastive learning framework developed for graphs with the aim of maximizing the mutual information between different augmented views of the same graph. It uses a noise-contrastive estimation approach to differentiate between positive and negative samples. Given a batch of $N$ graphs,  we re-annotate $\vz_1$, $\vz_2$ as $\vz_{i,1}$, $\vz_{i,2}$ for the $i$-th graph in the minibatch. Then the contrastive loss~\cite{zhu2021empirical} is:
\begin{equation}
\label{eq:graphcl}
\begin{aligned}
  \ell^\text{(graphcl)}_i
   = & -\frac{1}{2} \left( \log \frac{\exp(\text{sim}(\vz_{i,1}, \vz_{i,2})/\tau)}{\sum_{j = 1}^{N} \exp(\text{sim}(\vz_{i,1}, \vz_{j,2})/\tau)} \right. \\
   &  + \left. \log \frac{\exp(\text{sim}(\vz_{i,2}, \vz_{i,1})/\tau)}{\sum_{j = 1}^{N} \exp(\text{sim}(\vz_{i,2}, \vz_{j,1})/\tau)} \right)
\end{aligned}
\end{equation}
where $\tau$ denotes the temperature parameter, $\text{sim}(\cdot, \cdot)$ is the cosine similarity function.  The final loss is computed across all positive pairs in the batch.

\noindent \textbf{SimSiam} \cite{chen2021exploring} learns representations by maximizing the similarity between differently augmented views of the same sample. Unlike GraphCL, it doesn't rely on negative samples. The SimSiam framework processes two augmented views of a graph through the same encoder network. After encoding, SimSiam applies an MLP predictor to one view and a stop-gradient operation to the other view. The model then maximizes the similarity between these two processed views.
Specifically, for two augmented views of a graph, we have:
\begin{equation}
      \vp_1 = \text{MLP}(\vz_1) \quad \quad 
       \vp_2 = \text{MLP}(\vz_2),
\end{equation}
where $\vz_1$ and $\vz_2$ are the encoded graph representations of the two views. $\vp_1$ and $\vp_2$ are their respective predictions after passing through the MLP. 
The objective is to minimize their negative cosine similarity:
\begin{equation}
\begin{aligned}
    & \mathcal{D}(\vp_1, \text{stopgrad}(\vz_2)) = - \frac{\vp_1}{\| \vp_1 \|_2} \cdot \frac{\text{stopgrad}(\vz_2)}{\| \text{stopgrad}(\vz_2) \|_2} \\
    & \mathcal{D}(\vp_2, \text{stopgrad}(\vz_1)) = - \frac{\vp_2}{\| \vp_2 \|_2} \cdot \frac{\text{stopgrad}(\vz_1)}{\| \text{stopgrad}(\vz_1) \|_2}.
\end{aligned}
\end{equation}
The stop-gradient (stopgrad) operation is a key component of SimSiam. It blocks the gradients of $\text{stopgrad}(\vz_1)$ and $\text{stopgrad}(\vz_2)$ during backpropagation, treating these terms as constants when computing gradients. This process prevents direct optimization of the encoder through these paths, which is crucial for avoiding trivial solutions and encouraging the model to learn meaningful representations. The objective function is implemented as follows:
\begin{equation}
\ell^\text{(simsiam)} = \frac{1}{2} \mathcal{D}(\vp_1, \text{stopgrad}(\vz_2)) + \frac{1}{2} \mathcal{D}(\vp_2, \text{stopgrad}(\vz_1)).
\label{eq:simsiamloss}
\end{equation}

\section{Theoretical Analysis}
\label{sec:theory}
To investigate the importance of preserving semantics in augmented graphs for GRL, in this section, we theoretically analyze the errors of classifying graph embeddings produced by (1) an encoder trained with semantics-preserving augmentations, denoted by $f_\mathrm{enc}^{\mathrm{sp}}(\cdot)$; and (2) an encoder trained with semantics-agnostic augmentations, denoted by $f_\mathrm{enc}^{\mathrm{sa}}(\cdot)$. Our main theorem (Theorem \ref{th:1}) shows that in certain classification scenarios, the error under $f_\mathrm{enc}^{\mathrm{sp}}(\cdot)$ is close to zero, whereas for $f_\mathrm{enc}^{\mathrm{sa}}(\cdot)$, the error is close to $\frac{1}{2}$, which is equivalent to random guessing.

\begin{figure}[h]
    \centering
    \includegraphics[width=0.7\linewidth]{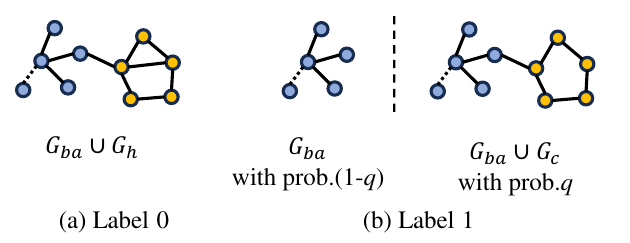}
    \vspace{-0.1cm}
    \caption{Exemplar graphs in modified BA-2motifs.}
    \label{fig:mba2motif}
\end{figure}
To demonstrate our analysis, we modify the widely studied benchmark BA2-Motifs \cite{luo2020parameterized}. As shown in Fig.~\ref{fig:mba2motif}, the dataset has two classes of graphs. The first class consists of graphs generated by taking the union of a Barabási-Albert (BA) graph \cite{albert2002statistical} and a house motif. The second class also has a BA graph as the base, which is optionally attached to a cycle motif with probability $q\in (0,1)$. Formally, let $P_{G_{ba}}$ be a probability distribution of BA graphs, let $G_{h}$ represent the house motif subgraph, with five nodes and six edges, and let $G_{c}$ represent the cycle motif with five nodes and five edges. Graphs with label 0 are of the form $G_{ba}\cup G_{h}$, {\em i.e.}, attach $G_{h}$ to $G_{ba}$, where $G_{ba}\sim P_{G_{ba}}$. Graphs with label 1 can either be $G_{ba}$ or $G_{ba}\cup G_{c}$, with probabilities $1-q$ and $q$, respectively.

We consider two types of edge-dropping-based augmentations as follows.
\begin{itemize}[leftmargin=*]
    \item \textbf{Semantics-Agnostic Augmentation ($P^{sa}_{G'|G}$).} Each edge in a graph is dropped independently with a probability $p\in (0,1)$.
        \item \textbf{Semantics-Preserving Augmentation ($P^{sp}_{G'|G}$).} Each edge in the BA graph $G_{ba}$ and the cycle motif $G_c$ is dropped with probability $p$, but the edges of the house motif $G_h$ are left unchanged.
\end{itemize}

Next, we use {\em empirical contrastive learner} (ECL), which is a generalized notion of the aforementioned encoders $f_\mathrm{enc}^{\mathrm{sp}}$ and $f_\mathrm{enc}^{\mathrm{sa}}$, to analyze the downstream classification errors using their output embeddings. Note that the proposed \ours-GRL method is an instance of the $f_\mathrm{enc}^{\mathrm{sp}}$ encoder.

An ECL aims to learn similar embeddings of graphs that are deemed similar by some selected distance measure. Equivalently, suppose the graphs can be partitioned into at most $\kappa=2$ clusters based on a distance measure $d_c$, the goal of an ECL is to learn graph embeddings such that graphs within the same cluster have similar embeddings. To quantify the similarity of graph pairs, we define a distance measure $d_c(G,G')$ as the absolute difference in the number of cycles in graphs $G$ and $G'$. For instance, $d_c(G_c,G_h)=2$ since the cycle motif has one cycle and the house motif has three cycles. With this distance measure as an instance, the following theorem, whose proof is deferred to Appendix A
, shows that under semantics-preserving augmentation $P^{sp}_{G'|G}$, the ECL can successfully learn graph embeddings such that the same labeled graphs belong to the same cluster. Using such embeddings for classification leads to zero classification error. In contrast, under semantics-agnostic augmentation $P^{sa}_{G'|G}$, the ECL fails to learn graph embeddings with such a property, leading to an error rate close to $\frac{q}{2}$.

\begin{Theorem} 
\label{th:1}
In the modified BA-2Motifs classification task described above, consider the ECLs $f_\text{enc}^{sa}$ and $f_\text{enc}^{sp}$, which correspond to the semantic-agnostic augmentation $P^{sa}_{G'|G}$ and the semantic-preserving augmentation $P^{sp}_{G'|G}$, respectively, 
with edge-drop probability $p > 0.3$, and as the size of the unlabeled training set $\mathcal{T}_u$ grows asymptotically large, the following hold:
\\i) The error rate of an empirical risk minimization (ERM) operating on the embeddings from $f_\mathrm{enc}^{\mathrm{sa}}(G)$ converges to $\dfrac{q}{2}$.
\\ii) The error rate of an ERM operating on the embeddings from $f_\mathrm{enc}^{\mathrm{sp}}(G)$ converges to $0$. 
\end{Theorem}
This theorem suggests that semantics-preserving augmentations can endow the ECL with the capability of producing embeddings that facilitate perfect classification in the asymptotic regime, while semantics-agnostic augmentations inevitably lead to significantly higher error rates. Although the theorem is demonstrated on a benchmark dataset, the underlying principles are potentially extendable to more real-world scenarios, where critical substructures determine graph labels~\cite{ying2019gnnexplainer,yuan2021explainability}. 

\begin{table*}[t!]
    \centering
        \caption{Comparison of different graph augmentation methods using GraphCL as the GRL framework.}
    \vspace{-0.2cm}
    \begin{tabular}{l|l|cccccc}
    \hline
    \multicolumn{2}{c|}{Augmentation Method} & {\mutag} & {\benz} & Alkane-Car.  & Fluoride-Car. & {\dd} & {\prot}  \\ \hline
    \multirow{2}{*}{Node Dropping} & Vanilla  & 0.803$_{\pm 0.030}$ & \textbf{0.767$_{\pm 0.049}$} & $0.965_{\pm 0.038}$ & 0.648$_{\pm 0.067}$ & \textbf{0.653}$_{\pm 0.070}$ & 0.728$_{\pm 0.073}$\\ 
                           & {\ours}    & \textbf{0.860$_{\pm 0.020}$} & 0.765$_{\pm 0.050}$ & \textbf{0.979}$_{\pm 0.020}$ & \textbf{0.656$_{\pm 0.066}$} & 0.649$_{\pm 0.065}$ & \textbf{0.744$_{\pm 0.077}$}\\ 
    \hline
    \multirow{2}{*}{Edge Dropping } & Vanilla & 0.858$_{\pm 0.027}$ & 0.753$_{\pm 0.043}$ & 0.942$_{\pm 0.047}$ & 0.659$_{\pm 0.044}$ & 0.660$_{\pm 0.048}$ & 0.702$_{\pm 0.077}$ \\
                    & {\ours} & \textbf{0.861$_{\pm 0.053}$} & \textbf{0.754}$_{\pm 0.050}$ & \textbf{0.948}$_{\pm 0.026}$ & \textbf{0.662$_{\pm 0.053}$} & \textbf{0.665}$_{\pm 0.050}$ & \textbf{0.757$_{\pm 0.055}$}  \\  
    \hline
    \multirow{2}{*}{Attribute Masking} & Vanilla  & 0.820$_{\pm 0.064}$  & \textbf{0.762$_{\pm 0.052}$} & 0.967$_{\pm 0.032}$ & 0.653$_{\pm 0.071}$ & 0.616$_{\pm 0.060}$ &  0.683$_{\pm 0.077}$\\ 
                                   & {\ours}    & \textbf{0.850$_{\pm 0.047}$} & 0.750$_{\pm 0.032}$ & \textbf{0.975}$_{\pm 0.027}$ & \textbf{0.658$_{\pm 0.046}$} & \textbf{0.624}$_{\pm 0.048}$ & \textbf{0.715$_{\pm 0.057}$} \\ 
    \hline
    \multirow{2}{*}{Subgraph} & Vanilla & 0.842$_{\pm 0.038}$ & 0.762$_{\pm 0.034}$ & 0.973$_{\pm 0.027}$ & \textbf{0.655$_{\pm 0.044}$} & 0.651$_{\pm 0.060}$ & 0.704$_{\pm 0.077}$    \\ 
                                   & {\ours}  & \textbf{0.846$_{\pm 0.037}$} & \textbf{0.765$_{\pm 0.056}$} & \textbf{0.987}$_{\pm 0.019}$ & 0.644$_{\pm 0.059}$ & \textbf{0.663}$_{\pm 0.065}$ & \textbf{0.728$_{\pm 0.082}$} \\ 
    \hline
    \multirow{2}{*}{Mixup} & Vanilla & 0.850$_{\pm 0.024}$ & 0.766$_{\pm 0.040}$ & 0.971$_{\pm 0.019}$ & 0.650$_{\pm 0.050}$ & \textbf{0.643}$_{\pm 0.040}$ & 0.728$_{\pm 0.072}$ \\ 
                                   & {\ours}  & \textbf{0.852}$_{\pm 0.027}$ & \textbf{0.769$_{\pm 0.061}$} & \textbf{0.975}$_{\pm 0.020}$ & \textbf{0.661$_{\pm 0.060}$} & 0.640$_{\pm 0.032}$ & \textbf{0.750$_{\pm 0.066}$} \\ 
    \hline
    \end{tabular}
    \label{tab:aucresults_graphcl}
\end{table*}
\section{Experiments}
In this section, we conduct extensive experiments to evaluate the proposed \ours-GRL 
and compare it with the widely used augmentation methods and the state-of-the-art (SOTA) GRL methods.

\subsection{Experimental Setup}
\textbf{Datasets.} To evaluate the performance of \ours-GRL, we use eight benchmark real-world datasets with graph-level labels, including {\mutag} \cite{luo2020parameterized}, {\benz} \cite{agarwal2023evaluating}, {\alk} \cite{agarwal2023evaluating}, {\fluo} \cite{agarwal2023evaluating}, {\dd} \cite{dobson2003distinguishing}, and {\prot} \cite{dobson2003distinguishing, borgwardt2005protein}. Among them, \mutag, \benz, \alk\, and \fluo\ also provide the ground truth subgraphs that explain the classification of every graph instance, {\em i.e.}, the semantics pattern. This information will be used for our analysis of semantics-preservation. 
Detailed statistics and descriptions of the datasets can be found in Appendix C.1.

\noindent{\textbf{Baselines.}} The key contribution of this work is a novel graph augmentation method \ours, which is agnostic to the choice of the GRL method. Therefore, in the experiment, we consider two widely used contrastive learning algorithms, {\em i.e.}, GraphCL \cite{graphcl} and SimSiam \cite{chen2021exploring}, as the basic GRL method on augmented graphs, and compare \ours\ with its semantics-agnostic counterparts (``Vanilla''): 
(1) {\em Node-Dropping}; (2) {\em Edge-Dropping}; (3) {\em Attribute-Masking}; (4) {\em Subgraph-Sample}; and (5) {\em Mixup} .

Moreover, we compare our \ours\ enhanced GRL, namely \ours-GRL, with other SOTA approaches for GRL, the detailed configurations and results of these compared GRL methods can be found in Appendix C.2 and Appendix D.6, respectively.

\noindent{\textbf{Implementation.}} We evaluate the classification performance based on the learned embeddings provided by different GRL methods. Specifically, following \cite{graphcl}, after training a GNN by a GRL algorithm, the embeddings generated by the GNN are fed to an SVM for classification. Detailed information regarding the implementation of the experiment is delineated in Appendix C.3.
\vspace{-0.1cm}
\subsection{Performance Analysis of Augmentation Methods}

We report the mean accuracy of graph classification over 10 random runs. 
Table \ref{tab:aucresults_graphcl} summarizes the results of comparing \ours\ with different augmentation techniques using GraphCL as the GRL framework. 
From the results, we have several observations.
First, different augmentation techniques may be useful to different extents on different datasets, where the methods modifying graph structures ({\em e.g.}, Subgraph, Mixup, Edge Dropping, etc.) are generally better than Attribute Masking. This is because structural changes imply more variances on graphs than node features, which is a desideratum for augmenting the dataset. 
Second, in most cases, our method \ours\ outperform each Vanilla augmentation technique, with up to 7.83\% relative improvement (Edge-Dropping on \prot) using GraphCL, indicating its generalizability across various augmentations and datasets, and its GRL-agnositic design for plug-and-play with different GRL methods. Finally, \ours\ is less sensitive to the loss of semantics caused by random perturbations, such as the degraded accuracy of Vanilla Node Dropping on \mutag\ in Table \ref{tab:aucresults_graphcl}. This is because the perturbation is constrained to the marginal subgraphs outside the semantic patterns by \ours, leading to its robustness to the potentially arbitrary changes caused by perturbations.

Moreover, Appendix D.1 presents a comparison of \ours\ with different augmentation techniques under the SimSiam GRL framework. 
To enrich the baseline comparisons, we also incorporate the NodeSam augmentation \cite{yoo2022model} in Appendix D.2. 
Further, We evaluate its performance with Graph Isomorphism Network (GIN) \cite{xuhow2019} as another base GNN model in Appendix D.3. 
Finally, we adopt Refine \cite{wang2021towards} as the explainer, with results reported in Appendix D.4. 
These experiments demonstrate the generalizability of \ours\ across different datasets, GNN architectures, augmentation methods, and explainers.
\vspace{-0.12cm}

\subsection{Experiments on Synthetic and Large Datasets}
To further demonstrate the robustness of {\ours} on different types of data, we conduct experiments with the GraphCL framework on a synthetic dataset {\bamo}~\cite{luo2020parameterized}, a real-world dataset {\hiv} ~\cite{luong2024fragment} with a large number of graphs, and a real-world dataset {\RB}~\cite{yanardag2015deep} with large scale graphs. 
The results are shown in Table~\ref{tab:aucresults_graphcl_appendix}. We find that our method achieves the best results across almost all data augmentation methods. Specifically, on the {\bamo} dataset, our method improves by an average of 17.3\% compared to Vanilla. Note that, since the node features in {\bamo} have only one dimension, the feature masking augmentation method cannot be applied.

\begin{table}[t]
    \setlength{\tabcolsep}{2.3pt}
    \small
    \centering
    \caption{Comparison of different graph augmentation methods on synthetic and large datasets.}
    \vspace{-0.2cm}
    \begin{tabular}{l|l|ccc}
    \hline
    \multicolumn{2}{c|}{Augmentation Method} & {\bamo} & {\hiv}  & REDDIT-B.  \\ \hline
    \multirow{2}{*}{Node Dropping} & Vanilla & 0.739$_{\pm 0.126}$  & 0.634$_{\pm 0.032}$ & 0.779$_{\pm 0.024}$ \\ 
                               & {\ours}     & \textbf{0.874$_{\pm 0.148}$}  & \textbf{0.643}$_{\pm 0.028}$ & \textbf{0.784}$_{\pm 0.039}$ \\ 
    \hline
    \multirow{2}{*}{Edge Dropping} & Vanilla & 0.603$_{\pm 0.127}$ & \textbf{0.640}$_{\pm 0.034}$  & 0.782$_{\pm 0.045}$ \\
                    & {\ours} & \textbf{0.701$_{\pm 0.197}$}  & 0.638$_{\pm 0.030}$  & \textbf{0.787}$_{\pm 0.031}$ \\  
    \hline
    \multirow{2}{*}{Attribute Masking} & Vanilla & -  & 0.621$_{\pm0.017 }$ & 0.816$_{\pm 0.022}$ \\ 
                                   & {\ours}     & -  & \textbf{0.632}$_{\pm 0.015}$ & \textbf{0.825}$_{\pm 0.026}$ \\ 
    \hline
    \multirow{2}{*}{Subgraph} & Vanilla & 0.781$_{\pm 0.145}$  & 0.638$_{\pm 0.033}$  & 0.779$_{\pm 0.036}$  \\ 
                                   & {\ours}     & \textbf{0.786$_{\pm 0.180}$}  & \textbf{0.640}$_{\pm 0.018}$ & \textbf{0.786}$_{\pm 0.035}$ \\ 
    \hline
    \multirow{2}{*}{Mixup} & Vanilla & 0.691$_{\pm 0.164}$  & 0.632$_{\pm 0019}$ & 0.777$_{\pm 0.026}$ \\ 
                                   & {\ours}     & \textbf{0.694$_{\pm 0.143}$}  & \textbf{0.642}$_{\pm 0.016}$ & \textbf{0.784}$_{\pm 0.025}$\\ 
    \bottomrule
    \end{tabular}
    \label{tab:aucresults_graphcl_appendix}
\end{table}

\subsection{Ablation Study}
\begin{figure}[h]
    \centering
    \includegraphics[width=1.0\linewidth]{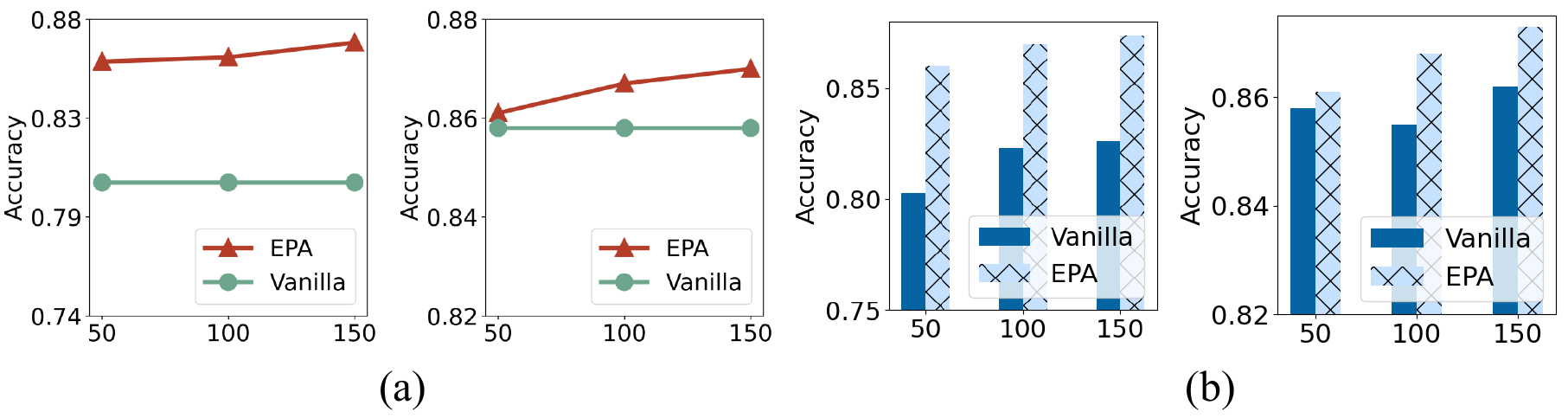}
    \vspace{-0.5cm}
    \caption{Accuracy with different numbers of (a) \# pre-training samples and (b) \# downstream training samples.}
    \label{fig:parameters}
\end{figure}

In this section, we use \mutag\ dataset for ablation study. To understand \ours's label-efficiency, we change the number of labeled graphs for explainer pre-training, {\em i.e.}, $|\gT_\ell|$, within \{50, 100, 150\} while fixing the number of training graphs for SVM as 50. Fig. \ref{fig:parameters}(a) summarizes the results of using Node/Edge Dropping as the base perturbation in \ours. Extensive ablation studies of other perturbation methods are deferred to Appendix D.7. From Fig. \ref{fig:parameters}(a), as $|\gT_\ell|$ increases, \ours-GRL gains higher accuracy with different perturbation methods. The ``Vanilla'' method has constant results because it is semantics-agnostic. As such, \ours\ is capable of making effective use of more labels for providing better augmentations. However, as labeling is usually expensive, we restrict most of our experiments to the challenging region with only 50 labeled graphs. Additionally, we evaluate the impacts of varying numbers of labeled graphs for the downstream training of SVM. As shown in Fig. \ref{fig:parameters}(b) (and Appendix D.7), both ``Vanilla'' and \ours\ augmented graphs enable better graph embeddings for effective downstream training with more labels. In particular, \ours-GRL consistently produce better embeddings than the baseline method as indicated by the clear accuracy gap.

\vspace{-0.3cm}

\subsection{Impact of Explanation Quality on Model Performance}
\begin{figure}[h]
    \centering
    \includegraphics[width=0.8\linewidth]{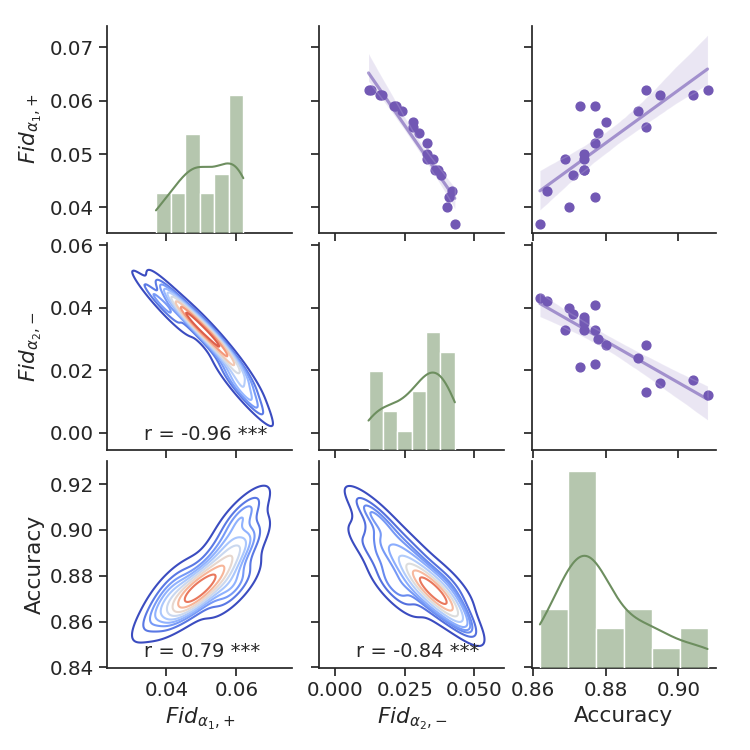}
    \vspace{-0.2cm}
    \caption{Correlation between $Fid_{\alpha_1,{+}}$, $Fid_{\alpha_2,{-}}$, and Accuracy on {\mutag} dataset. The value $r$ represents the Pearson correlation coefficient. Statistical significance is denoted by $\ast$, with $\ast\ast\ast$ indicating a p-value of $p \leq 0.01$ for testing non-correlation.}
    \label{fig:correlation}
\end{figure}

To analyze the relationship between explanation quality and model performance, we evaluate how the fidelity of explanation subgraphs influences classification accuracy. We employ two theoretically grounded fidelity metrics, $Fid_{\alpha_1,{+}}$ and $Fid_{\alpha_2,{-}}$, proposed by \cite{zheng2024robustfidelityevaluatingexplainability}. The detailed formulation of these metrics is provided in Appendix D.8. A higher $Fid_{\alpha_1,{+}}$ or a lower $Fid_{\alpha_2,{-}}$ score indicates better explanation quality.

To systematically evaluate this relationship, we introduce controlled perturbations to the explanation subgraphs generated by our trained explainer, following the setting in~\cite{zheng2024robustfidelityevaluatingexplainability}. For each explanation subgraph, we randomly remove a proportion $\beta$ of its edges and replace them with the same number of randomly selected non-explanation edges, where $\beta$ ranges from 0 to 1 in increments of 0.05. Fig. \ref{fig:correlation} presents the correlation between both fidelity metrics and classification accuracy on the \mutag\ dataset. The strong correlation observed between the fidelity metrics and classification performance demonstrates that higher-quality explanations lead to better representation learning, validating our approach of preserving semantic structures for effective graph augmentation.

\subsection{Additional Analysis}
Appendix D.5 reports the training time; D.6 compares our method with GRL methods; D.9 investigates the impact of the subgraph size parameter $k$; D.10 presents case studies highlighting EPA's semantic preservation; and D.11 visualizes \ours\ embeddings on various datasets.

\section{Conclusion}
In this paper, we study data augmentation methods for graph contrastive learning. In contrast to most of the existing methods, which focus on structural perturbations but overlook the importance of preserving semantic information, we propose a novel framework {\ours-GRL} to deal with both. {\ours-GRL} incorporates the explanatory patterns of a graph into its data augmentations by leveraging a few labeled graphs and trains a GRL model with more unlabeld graphs, establishing a semi-supervised learning paradigm. We perform theoretical analysis and conduct comprehensive experiments. The results validate the effectiveness of the proposed method. 

\section{Acknowledgments}
This project was partially supported by NSF grants IIS-2529283, ECCS-2242700 and CCF-2241057. The views
and conclusions contained in this paper are those of the authors and should not be interpreted as
representing any funding agencies.

\bibliography{aaai2026}

\newpage
\appendix
\onecolumn
\section{A. Detailed Theoretical Analysis}
\label{sec:app:proof}
In this section, we provide the formal definitions and proofs that support our theoretical analysis presented in Sec.~\ref{sec:theory}.

\subsection{1. Definition of the Empirical Contrastive Learner}
At a high level, contrastive learning models cluster the set of all possible input graphs into positive and negative pairs,  and consequently train a network such that the embeddings of positive pairs are as similar as possible while ensuring that negative pairs are assigned maximally discriminative embeddings. Due to the nature of augmentations, different input graphs can yield augmented graphs that are close or similar to each other, resulting in a situation where a given graph could simultaneously be categorized as both positive and negative relative to another graph. It is tempting to define the ECL to empirically count the number of times two graphs are paired as positives and the number of times they are paired as negatives, assigning identical or discriminative embeddings based on which count is greater. However, this introduces a circular problem: for example, if graph $G$ is more positively paired than negatively paired with both graphs $G'$ and $G''$, then we assign similar embeddings to all three. However, if  $G'$ and $G''$ are more negatively paired than positively paired, this assignment becomes inconsistent. Consequently, we define the ECL as an algorithm that considers all possible partitions of the set of graphs. Each partition is assigned a score based on the number of positive pairs within elements of the same subset of the partition, minus the number of negative pairs. The ECL then outputs embeddings based on the highest-scoring partition, with elements of the same subset receiving similar embeddings. 

Given a training set\footnote{Note that in practice, one could also reuse the labeled training set elements $\mathcal{T}_\ell$ in training the contrastive learning model.} $\mathcal{T}_u=\{G_i, i\in [n]\}$ and augmentation mapping $P_{G'|G}$, the augmented set $\mathcal{A}=\{(G_{i,1},G_{i,2}), i\in [n]\}$ consists of pairs $(G_{i,1},G_{i,2})$ generated independently based on $P_{G'|G}(\cdot|G_i)$. We define a distance measure $d(G,G')\in \mathbb{R}_{\geq 0}$ between pairs of graphs.  The ECL assigns similar embeddings to graphs whose augmentations have a small distance. This is formalized as below. 
\begin{Definition}[\textbf{Empirical Contrastive Learner}]
Let  $P_{G'|G}$ be an augmentation mapping, $d(\cdot,\cdot)$ be a distance measure, and $\kappa>0$ be a clustering coefficient. For a threshold $\epsilon>0$, and an arbitrary pair of graphs $G,G'$, their pairwise score is defined as:
\begin{align*}
    s_{\epsilon}(G,G')= &|\{G_{i}\in \mathcal{T}_u| d(G,G_{i,1})<\epsilon,  d(G',G_{i,2})<\epsilon\}|
  \\&-|\{G_{i}\in \mathcal{T}_u| d(G,G_{i,1})<\epsilon,  d(G',G_{i,2})>\epsilon\}|,
\end{align*}
and their symmetric score is $\overline{s}_{\epsilon}(G,G')= s_{\epsilon}(G,G')+ s_{\epsilon}(G',G)$. Let $\mathsf{P}$ denote the set of all possible partitions of $\mathcal{T}_u$. Then, for $\mathcal{P}\in \mathsf{P}$, let the partition elements corresponding to  $\mathcal{P}$ be denoted by $\mathcal{G}_1,\mathcal{G}_2,\cdots,\mathcal{G}_{|\mathcal{P}|}$. The partition score is defined as:
\begin{align*}
    \pi_{\epsilon}(\mathcal{P})=
    \sum_{i=1}^{|\mathcal{P}|}\sum_{G,G'\in \mathcal{G}_i} \overline{s}_{\epsilon}(G,G'). 
\end{align*}
The empirically optimal $\kappa$-partition $\mathcal{P}^*$ is defined as the one with the highest partition score among all elements of $\mathsf{P}$ with size at most $\kappa$. Then,  the ECL is defined as follows:  \begin{align*}
f_\mathrm{enc}(G)= \mathbf{Z}_{\mathcal{G}^*(G)}, \quad \mathcal{G}^*(G)= \argmax_{\mathcal{G}^*_i\in \mathcal{P}^*} \frac{1}{|\mathcal{G}^*_i|}\sum_{G'\in \mathcal{G}^*_i} \overline{s}_{\epsilon}(G,G'),
\end{align*}
where $\mathbf{Z}_{\mathcal{G}^*}, \mathcal{G}^*\in \mathcal{P}^*$ are maximally distinct and normalized vectors\footnote{A collection of vectors $\mathbf{z}_i, i\in [m]$ are called maximally distinct and normalized if each vector is normalized, i.e., $|\mathbf{z}_i| = 1$ for all $i$, and the minimum pairwise Euclidean distance between any two distinct vectors is maximized.}. 
\end{Definition}

The ECL assigns embeddings to graphs based on the frequency of positive and negative pairings induced by the augmentation process. It seeks a partition of the graph dataset that maximizes the alignment of positive pairs within the same cluster while separating negative pairs into different clusters.

\subsection{2. Refined Statement and Proof of Theorem~\ref{th:1}}
Building upon the formal definition of the ECL, we now present a refined statement of Theorem~\ref{th:1}, providing precise conditions and outcomes.
\begin{Theorem}
\label{th:refine1}
In the modified BA2-Motifs classification task described in the prequel,
consider the pair of ECLs $f_\text{enc}^{sa}$ and $f_\text{enc}^{sp}$ characterized by $(P^{sa}_{G'|G},d_c(G,G'), \kappa,\epsilon)$ and
$(P^{sp}_{G'|G},d_c(G,G'), \kappa,\epsilon)$, respectively, with edge-drop probability $p>0.3$, clustering coefficient $\kappa=2$, and threshold $\epsilon<1$. 
As the size of the training set $\mathcal{T}_u$ grows asymptotically large, the following hold:
\\i) The error rate of an ERM operating on $f_\mathrm{enc}^{\mathrm{sa}}(G)$ converges to $\frac{q}{2}$.
\\ ii)The error rate of an ERM operating on  $f_\mathrm{enc}^{\mathrm{sp}}(G)$ converges to 0.
\end{Theorem}

\subsection{3. Proof of Theorem \ref{th:refine1}}
We note that the application of edge-drop augmentation to the two classes of graphs yields a graph with either $0,1$ or $3$ cycles. The reason is that the BA graph itself does not have any cycles, and the house motif originally has $3$ cycles, which may produce  $0,1$ or $3$  cycles after edge-drop, and the cycle motif orignally has $1$ cycle  and may produce $0$ or $1$ cycle after edge drop. Let $\mathcal{C}_i, i\in \{0,1,3\}$ represent the collection of all graphs with $i$ cycles. Then, there are four possible partitions with size less than or equal to $\kappa=2$:
\begin{align*}
   & \mathcal{P}_1=\{\{\mathcal{C}_0,\mathcal{C}_1,\mathcal{C}_3\}\},
\\&   \mathcal{P}_2=\{\{\mathcal{C}_0,\mathcal{C}_1\},\{\mathcal{C}_3\}\},
\\& \mathcal{P}_3=\{\{\mathcal{C}_0\},\{\mathcal{C}_1,\mathcal{C}_3\}\},
\\& 
\mathcal{P}_4=\{\{\mathcal{C}_1\},\{\mathcal{C}_0,\mathcal{C}_3\}\}.
\end{align*}
We evaluate the partition score for each of these partitions under the $P^{sa}_{G'|G}$ and $P^{sp}_{G'|G}$ augmentations, to find the highest scoring partition produced by the corresponding ECL, which in turn determined the ERM accuracy. 
\\\textbf{Case 1: Semantic-Agnostic Augmentations.}
Recall that $\mathcal{T}_u=\{G_i| i\in [|\mathcal{T}_u|]\}$ represents the training set and $\mathcal{A}=\{(G_{i,1},G_{i,2}|i \in [|\mathcal{T}_u|]\}$ represents the augmented pair of graphs. Let $\omega_{k,\ell}, k,\ell \in\{0,1,3\}$ be the fraction of augmented pairs $(G_{i,1},G_{i,2})$ with $k$ and $\ell$ cycles, respectively. Then, 
\begin{align*}
&G,G'\in \mathcal{C}_0 \Rightarrow \overline{S}_{0,0}\triangleq \overline{s}_{\epsilon}(G,G')= 2|\mathcal{T}_u|(\omega_{0,0}-\omega_{0,1}-\omega_{0,3})
\\& 
G,G'\in \mathcal{C}_1 \Rightarrow \overline{S}_{1,1}\triangleq \overline{s}_{\epsilon}(G,G')= 2|\mathcal{T}_u|(\omega_{1,1}-\omega_{0,1}-\omega_{1,3})
\\&
G,G'\in \mathcal{C}_3 \Rightarrow \overline{S}_{3,3}\triangleq \overline{s}_{\epsilon}(G,G')= 2|\mathcal{T}_u|(\omega_{3,3}-\omega_{0,3}-\omega_{1,3})
\\&
G\in \mathcal{C}_0,G'\in \mathcal{C}_1 \Rightarrow
\overline{S}_{0,1}\triangleq \overline{s}_{\epsilon}(G,G') 
\\& \quad \quad \quad \quad \quad \quad \quad = |\mathcal{T}_u|(2\omega_{0,1}-\omega_{0,0}-\omega_{0,3}-\omega_{1,1}-\omega_{1,3})
\\&
G\in \mathcal{C}_0,G'\in \mathcal{C}_3
\Rightarrow 
\overline{S}_{0,3}\triangleq
\overline{s}_{\epsilon}(G,G')
\\& \quad \quad \quad \quad \quad \quad \quad = |\mathcal{T}_u|(2\omega_{0,3}-\omega_{0,0}-\omega_{0,1}-\omega_{1,3}-\omega_{3,3})
\\&
G\in \mathcal{C}_1,G'\in \mathcal{C}_3 \Rightarrow 
\overline{S}_{1,3}\triangleq
\overline{s}_{\epsilon}(G,G')
\\& \quad \quad \quad \quad \quad \quad \quad= |\mathcal{T}_u|(2\omega_{1,3}-\omega_{0,1}-\omega_{1,1}-\omega_{0,3}-\omega_{3,3})
\end{align*}
Furthermore:
\begin{align*}
&\pi_{\mathcal{P}_1}=2(\overline{S}_{0,0}+\overline{S}_{0,1}+\overline{S}_{0,3}+\overline{S}_{1,1}+\overline{S}_{1,3}+\overline{S}_{3,3}),
\\& \pi_{\mathcal{P}_2}= 2(\overline{S}_{0,0}+\overline{S}_{0,1}+\overline{S}_{1,1}+\overline{S}_{3,3})
    \\&
    \pi_{\mathcal{P}_3}= 2(\overline{S}_{0,0}+\overline{S}_{1,1}+\overline{S}_{1,3}+\overline{S}_{3,3}),
    \\&
    \pi_{\mathcal{P}_4}= 2(\overline{S}_{1,1}+\overline{S}_{0,0}+\overline{S}_{0,3}+\overline{S}_{3,3}).
\end{align*}
Note that the term $\overline{S}_{0,0}+\overline{S}_{1,1}+\overline{S}_{3,3}$ is shared among all the partition scores and does not affect the choice of optimal partition. So, we focus on $\overline{S}_{0,1},\overline{S}_{0,3}$ and $\overline{S}_{1,3}$ terms. Following standard combinatorial arguments, we note that:
\begin{align*}
&\mathbb{E}(\omega_{0,0})=  \frac{1-q}{2} + \frac{q}{2}(1-(1-p)^5)^2 +\frac{1}{2}(p(1-(1-p)^5)
    \\& \quad \quad \quad \quad  \quad +(1-p)(1-(1-p)^2)(1-(1-p)^3))^2
    \\& \mathbb{E}(\omega_{0,1})= \frac{q}{2}(1-(1-p)^5)(1-p)^5 
    \\& \quad \quad \quad \quad  \quad    + \frac{1}{2}(p((1-p)^5)+((1-p)^4)(1-(1-p)^2)
    \\&\quad \quad \quad \quad  \quad +((1-p)^3)(1-(1-p)^3))(p(1-(1-p)^5)
    \\&\quad \quad \quad \quad  \quad  +(1-p)(1-(1-p)^2)(1-(1-p)^3))
   \\& \mathbb{E}(\omega_{1,1})= \frac{q}{2}((1-p)^5)^2 + \frac{1}{2}(p((1-p)^5)+((1-p)^4)(1-(1-p)^2)
   \\&\quad \quad \quad \quad  \quad +((1-p)^3)(1-(1-p)^3))^2
   \\& \mathbb{E}(\omega_{1,3})= \frac{1}{2}(p((1-p)^5)+((1-p)^4)(1-(1-p)^2)
   \\&\quad \quad \quad \quad  \quad +((1-p)^3)(1-(1-p)^3))((1-p)^6)
   \\& \mathbb{E}(\omega_{0,3})=\frac{1}{2}((1-p)^6)(p(1-(1-p)^5)
   \\&\quad \quad \quad \quad  \quad +(1-p)(1-(1-p)^2)(1-(1-p)^3))
   \\&\mathbb{E}(\omega_{3,3})= \frac{1}{2}((1-p)^6)^2
\end{align*}
Numerical evaluation yields $0>\mathbb{E}(\overline{S}_{1,3})> \max(\mathbb{E}(\overline{S}_{0,1}),\mathbb{E}(\overline{S}_{0,3}))$ for $p>0.3$. Consequently, using Heoffding's inequality, we conclude that $\mathcal{P}_{3}$ has the highest partition score among the four partitions $\mathcal{P}_i, i\in [4]$ with probability approaching one as $|\mathcal{T}_u|\to \infty$. As a result, graphs belonging to $\mathcal{C}_1$ and $\mathcal{C}_3$ are assigned the same embedding by the ECL. Consequently, the ERM is unable to distinguish between graphs with 1 and 3 cycles, e.g., graphs with house or cycle motifs. Thus, it achieves a probability of error at least equal to $\frac{q}{2}$. 
\\\textbf{Case 2: Semantic Preserving Augmentations.} Semantic preserving augmentations do not change the number of cycles in the input graph if the graph is attached to a house motif. Consequently, 
\begin{align*}
    \mathbb{E}(\omega_{0,1})>0,\quad  \mathbb{E}(\omega_{0,3})=\mathbb{E}(\omega_{1,3})=0.
\end{align*}
As a result, $0>\mathbb{E}(\overline{S}_{0,1})>\max(\mathbb{E}(\overline{S}_{0,3}),\mathbb{E}(\overline{S}_{1,3}))$. The partition $\mathcal{P}_{2}$ receives the highest score. So, graphs with 0 or 1 cycles are mapped to the same embedding and graphs with 3 cycles are mapped to a different embedding. Thus, an ERM applied to the output of the ECL achieves zero error rate.

\begin{table*}[h]
    \centering
    \caption{Overview of explanation-preserving data augmentations for graphs.}
    \begin{tabular}{c|c|c}
    \toprule
         \textbf{Data Augmentation} & \textbf{Type} & \textbf{Underlying Prior} \\ 
         \midrule
         Node Dropping & Nodes, edges & Missing inessential nodes does not affect semantics. \\
         Edge Dropping & Edges & The graph is robust to non-critical edge connectivity changes. \\
         Attribute Masking & Nodes & Loss of inessential attributes does not alter semantics. \\
         Subgraph & Nodes, edges & Local inessential sub-structures does not disrupt core semantics. \\
         Mixup & Nodes, edges & Mixing a semantic subgraph with another marginal subgraph preserves semantics. \\
         \bottomrule
    \end{tabular}
    \label{tab:augmentation}
    \end{table*}

\section{B. Detailed Algorithms}
\label{sec:app:algorithms}

\subsection{1. Explanation-Preserving Augmentation Algorithms}\label{app.augalg}
In this section, we provide the detailed pseudo code of the proposed explanation-preserving augmentation algorithms.  A summary is shown in Table~\ref{tab:augmentation}.

\noindent \textbf{Explanation-Preserving Node Dropping.} Algorithm \ref{alg:node_drop} outlines the process of explanation-preserving node dropping. Given an input graph $G$, a trained GNN explainer $\Psi(\cdot)$, and a dropping ratio $p$, the algorithm first extracts the explanation subgraph $G^{(\text{exp})}$ using the explainer. It then identifies the marginal subgraph $\Delta G$ by subtracting $G^{(\text{exp})}$ from $G$. Nodes in $\Delta G$ are sampled with probability $1-p$, and edges connecting these sampled nodes are retained. The algorithm constructs a sampled subgraph $G^{\text{(smp)}}$ from these nodes and edges. Finally, it combines $G^{(\text{exp})}$ and $G^{\text{(smp)}}$ to create the augmented graph $G'$. 

\begin{algorithm}[h]
\caption{Explanation-Preserving Node Dropping}
\begin{flushleft}
\textbf{Input:} Graph $G = (\mathcal{V}, \mathcal{E})$, trained GNN explainer $\Psi(\cdot)$, dropping ratio $p$ \\
\textbf{Output:} Explanation-preserving augmented Graph $G'$ \\
\begin{algorithmic}[1]
\State $G^{(\text{exp})} = \Psi(G)$ \Comment{Extract explanation subgraph}
\State $\Delta G = G \setminus G^{(\text{exp})}=(\Delta \mathcal{V}, \Delta \mathcal{E})$ \Comment{Compute marginal subgraph}
\State $\mathcal{V}^{\text{(smp)}} = \{v \mid v \in \Delta \mathcal{V}, \text{Bernoulli}(1-p) = 1\}$ \Comment{Sample nodes with probability $1-p$}
\State $\mathcal{E}^{\text{(smp)}} = \{(u, v) \mid (u, v) \in \Delta \mathcal{E}, u \in \mathcal{V}^{\text{(smp)}}, v \in \mathcal{V}^{\text{(smp)}}\}$ \Comment{Keep edges between sampled nodes}
\State $G^{\text{(smp)}} = (\mathcal{V}^{\text{(smp)}}, \mathcal{E}^{\text{(smp)}})$ \Comment{Construct sampled subgraph}
\State $G' = G^{(\text{exp})} \cup G^{\text{(smp)}}$ \Comment{Combine explanation and sampled subgraphs}
\State \textbf{return} $G'$
\end{algorithmic}
\end{flushleft}
\label{alg:node_drop}
\end{algorithm}

\noindent \textbf{Explanation-Preserving Edge Dropping.} Algorithm \ref{alg:edge_drop} describes the explanation-preserving edge dropping process. Similar to node dropping, it starts by extracting the explanation subgraph $G^{(\text{exp})}$ and identifying the marginal subgraph $\Delta G$. However, instead of sampling nodes, this algorithm samples edges from $\Delta G$ with probability $1-p$. It then collects all nodes that are incident to the sampled edges. The sampled subgraph $G^{\text{(smp)}}$ is constructed from these edges and their associated nodes. Finally, we combine $G^{(\text{exp})}$ and $G^{\text{(smp)}}$ to produce the augmented graph $G'$. 
Explanation-Preserving Node/Edge Dropping preserves the critical node/edge structure identified by the explainer while introducing variability in the connectivity of less important parts of the graph.

\begin{algorithm}[h]
\caption{Explanation-Preserving Edge Dropping}
\begin{flushleft}
\textbf{Input:} Graph $G = (\mathcal{V}, \mathcal{E})$, trained GNN explainer $\Psi(\cdot)$, dropping ratio $p$ \\
\textbf{Output:} Explanation-preserving augmented Graph $G'$ \\
\begin{algorithmic}[1]
\State $G^{(\text{exp})} = \Psi(G)$ \Comment{Extract explanation subgraph}
\State $\Delta G = G \setminus G^{(\text{exp})}= (\Delta \mathcal{V}, \Delta \mathcal{E} )$  \Comment{Compute marginal subgraph}
\State $\mathcal{E}^{\text{(smp)}} = \{e \mid e \in \Delta \mathcal{E}, \text{Bernoulli}(1-p) = 1\}$ \Comment{Sample edges with probability $1-p$}
\State $\mathcal{V}^{\text{(smp)}} = \{v \mid v \in \Delta \mathcal{V}, \exists e \in \mathcal{E}^{\text{(smp)}}, v \in e\}$ \Comment{Collect nodes of sampled edges}
\State $G^{\text{(smp)}} = (\mathcal{V}^{\text{(smp)}}, \mathcal{E}^{\text{(smp)}})$ \Comment{Construct sampled subgraph}
\State $G' = G^{(\text{exp})} \cup G^{\text{(smp)}}$ \Comment{Combine explanation and sampled subgraphs}
\State \textbf{return} $G'$
\end{algorithmic}
\end{flushleft}
\label{alg:edge_drop}
\end{algorithm}

\noindent \textbf{Explanation-Preserving Attribute Masking.} Algorithm \ref{alg:attribute_mask} presents the explanation-preserving attribute masking procedure. As with the previous methods, it begins by extracting the explanation subgraph $G^{(\text{exp})}$ and identifying the marginal subgraph $\Delta G$. The key difference lies in the treatment of node features. The algorithm generates a binary mask matrix $\mM^{\text{(smp)}}$ where each entry is sampled from a Bernoulli distribution with probability $1-p$. This mask is then applied to the feature matrix $\Delta \mX$ of the marginal subgraph through a pointwise product operation. The resulting $\mX^{\text{(smp)}}$ contains masked features for the non-critical nodes. The algorithm constructs $G^{\text{(smp)}}$ using the original structure of $\Delta G$ but with the masked features. Finally, it combines $G^{(\text{exp})}$ and $G^{\text{(smp)}}$ to create $G'$. This approach maintains the critical node attributes identified by the explainer while introducing controlled noise in the features of less important nodes.

\begin{algorithm}
\caption{Explanation-Preserving Attribute Masking}
\begin{flushleft}
\textbf{Input:} Graph $G = \{\mathcal{V}, \mathcal{E}, \mX\}$, trained GNN explainer $\Psi(\cdot)$, masking ratio $p$ \\
\textbf{Output:} Explanation-preserving augmented Graph $G'$ \\
\begin{algorithmic}[1]
\State $G^{(\text{exp})} = \Psi(G)$ \Comment{Extract explanation subgraph}
\State $\Delta G = G \setminus G^{(\text{exp})} = (\Delta \mathcal{V}, \Delta \mathcal{E}, \Delta \mX)$  \Comment{Compute marginal subgraph}
\State $\mM^{\text{(smp)}} \in \{0, 1\}^{|\Delta \mathcal{V}| \times d_n}, m^{\text{(smp)}}_{ij} \sim \text{Bernoulli}(1-p)$ \Comment{Generate binary mask matrix}
\State $\mX^{\text{(smp)}} = \Delta \mX \odot \mM^{\text(smp)}$ \Comment{Apply pointwise product}
\State $G^\text{(smp)} = (\Delta \mathcal{V}, \Delta \mathcal{E}, \mX^\text{(smp)})$ \Comment{Construct sampled subgraph}
\State $G' = G^{(\text{exp})} \cup G^\text{(smp)}$ \Comment{Combine explanation and sampled subgraphs}
\State \textbf{return} $G'$
\end{algorithmic}
\end{flushleft}
\label{alg:attribute_mask}
\end{algorithm}

\begin{algorithm}
\caption{Explanation-Preserving Subgraph}
\begin{flushleft}
\textbf{Input:} Graph $G = (\mathcal{V}, \mathcal{E})$, trained GNN explainer $\Psi(\cdot)$, sampling ratio $p$ \\
\textbf{Output:} Explanation-preserving augmented Graph $G'$ \\
\begin{algorithmic}[1]
\State $G^{(\text{exp})} = \Psi(G)$ \Comment{Extract explanation subgraph}
\State $\Delta G = G \setminus G^{(\text{exp})} = (\Delta \mathcal{V}, \Delta \mathcal{E})$  \Comment{Compute marginal subgraph}
\State $v_0 \gets$ Random node from $\Delta \mathcal{V}$
\State $\mathcal{V}^{\text{(smp)}} \gets {v_0}$
\State $\mathcal{V}^{\text{(neigh)}} \gets \mathcal{N}(v_0)$ \Comment{Get neighbors of $v_0$}
\While{$|\mathcal{V}^{\text{(smp)}}| < p |\Delta \mathcal{V}|$}
\State $v \gets$ Random node from $\mathcal{V}^{\text{(neigh)}} \setminus \mathcal{V}^{\text{(smp)}}$
\State $\mathcal{V}^{\text{(smp)}} \gets \mathcal{V}^{\text{(smp)}} \cup \{v\}$
\State $\mathcal{V}^{\text{(neigh)}} \gets \mathcal{V}^{\text{(neigh)}} \cup (\mathcal{N}(v) \setminus \mathcal{V}^{\text{(smp)}})$
\EndWhile
\State $\mathcal{E}^{\text{(smp)}} \gets \{(u, v) \mid (u, v) \in \Delta \mathcal{E}, u \in \mathcal{V}^{\text{(smp)}} \text{ or } v \in \mathcal{V}^{\text{(smp)}}\}$
\State $G^{\text{(smp)}} = (\mathcal{V}^{\text{(smp)}}, \mathcal{E}^{\text{(smp)}})$ \Comment{Construct sampled subgraph}
\State $G' = G^{(\text{exp})} \cup G^{\text{(smp)}}$ \Comment{Combine explanation and sampled subgraphs}
\State \textbf{return} $G'$
\end{algorithmic}
\end{flushleft}
\label{alg:subgraph}
\end{algorithm}

\noindent \textbf{Explanation-Preserving Subgraph.} Algorithm \ref{alg:subgraph} details the explanation preserving subgraph sampling process. After extracting the explanation subgraph $G^{(\text{exp})}$ and identifying the marginal subgraph $\Delta G$, the algorithm performs a random walk on $\Delta G$. It starts by randomly selecting an initial node $v_0$ from $\Delta G$. The algorithm then iteratively expands the sampled node set $\mathcal{V}^{\text{(smp)}}$ by randomly selecting nodes from the neighborhood of previously sampled nodes. This process continues until the desired number of nodes (determined by the sampling ratio $p$) is reached. The algorithm then collects all edges that have at least one endpoint in $\mathcal{V}^{\text{(smp)}}$ to form $\mathcal{E}^{\text{(smp)}}$. The sampled subgraph $G^{\text{(smp)}}$ is constructed from these nodes and edges. Finally, $G^{(\text{exp})}$ and $G^{\text{(smp)}}$ are combined to create the augmented graph $G'$. 

\noindent \textbf{Explanation-Preserving Mixup.} Algorithm \ref{alg:mixup} describes the explanation preserving mixup process. This method begins by extracting the explanation subgraph $G^{(\text{exp})}$ from the input graph $G$ and identifying its marginal subgraph $\Delta G$. It then randomly selects another graph $G'$ from the set of available graphs $\mathcal{G}$ and computes its marginal subgraph $\Delta G'$. The algorithm proceeds to mix these marginal subgraphs based on their relative sizes. If $\Delta G$ is smaller or equal in size to $\Delta G'$, it samples nodes from $\Delta G'$ to match $\Delta G$'s size and creates a one-to-one mapping between their nodes. If $\Delta G$ is larger, it samples a subset of $\Delta G$ to match $\Delta G'$'s size, maps these nodes, and retains the unmapped portion of $\Delta G$. The algorithm then constructs edges for the sampled nodes based on the original connections in $\Delta G'$. Finally, it combines the explanation subgraph $G^{(\text{exp})}$ with the mixed marginal subgraph to create the augmented graph $G^\text{(aug)}$.

\subsection{2. Graph Representation Learning with Explanation-Preserving Augmentations}
Algorithm \ref{alg:graphrl} presents the process of GraphRL with EPAs. We first initialize the GNN model $f_\text{pt}(\cdot)$, encoder $f_\text{enc}(\cdot)$, and projection head $g(\cdot)$. It then pre-trains the GNN model on the labeled dataset $\mathcal{T}_{\ell}$ for $e_w$ epochs using cross-entropy loss. After that, a parametric GNN explainer $\Psi(\cdot)$ is trained on $\mathcal{T}_{\ell}$. 
To train the encoder, for each epoch, we sample minibatches from both labeled and unlabeled datasets. For each graph in the minibatch, we apply two randomly selected EPA techniques from the set {EPA-Node Dropping, EPA-Edge Dropping, EPA-Attribute Masking, EPA-Subgraph, EPA-Mixup}. These augmentations generate two views of each graph while preserving their critical structures.
The algorithm then computes the self-supervised loss using either the contrastive loss or the SimSiam loss, depending on the chosen framework. Finally, it updates the encoder and projection head by minimizing this self-supervised loss.

\begin{algorithm}[h]
\caption{Explanation-Preserving Mixup}
\begin{flushleft}
\textbf{Input:} Graph $G = (\mathcal{V}, \mathcal{E})$, a set of graphs $\mathcal{G}$, trained GNN explainer $\Psi(\cdot)$ \\
\textbf{Output:} Explanation-preserving augmented Graph $G'$ \\
\begin{algorithmic}[1]
\State $G^{(\text{exp})} = \Psi(G)$ \Comment{Extract explanation subgraph from $G$}
\State $\Delta G = G \setminus G^{(\text{exp})} = (\Delta \mathcal{V}, \Delta \mathcal{E})$  \Comment{Compute marginal subgraph of $G$}
\State $\tilde{G} = (\tilde{\gV}, \tilde{\gE}) \gets$ Random graph from $\mathcal{G}$ \Comment{Sample a random graph}
\State $\Delta \tilde{G} = \tilde{G} - \tilde{G}^{(\text{exp})} = (\Delta \tilde{\gV}, \Delta \tilde{\gE})$ \Comment{Compute marginal subgraph of $\tilde{G}$}
\If{$|\Delta \mathcal{V}| \leq |\Delta \tilde{\gV}|$}
\State $\mathcal{V}^{(\text{smp})} \gets$ Random sample of $|\Delta \mathcal{V}|$ nodes from $\Delta \tilde{\gV}$
\State $f_\text{map}: \Delta \mathcal{V} \to \mathcal{V}^{(\text{smp})}$ \Comment{One-to-one mapping}
\State $\mathcal{E}^{(\text{smp})} = \{(f(u), f(v)) \mid (u, v) \in \Delta \tilde{\gE}, u, v \in \mathcal{V}^{(\text{smp})}\}$
\State $G^{(\text{smp})} = (\mathcal{V}^{(\text{smp})}, \mathcal{E}^{(\text{smp})})$
\Else
\State $\mathcal{V}^{(\text{mix})} \gets$ Random sample of $|\Delta \tilde{\gV}|$ nodes from $\Delta \mathcal{V}$
\State $f_\text{map}: \mathcal{V}^{(\text{mix})} \to \Delta \tilde{\gV}$ \Comment{One-to-one mapping for subset}
\State $\mathcal{E}^{(\text{mix})} = \{(f(u), f(v)) \mid (u, v) \in \Delta \tilde{\gE}, u, v \in \mathcal{V}^{(\text{mix})}\}$
\State $\mathcal{V}^{(\text{unmix})} = \Delta \mathcal{V} - \mathcal{V}^{(\text{mix})}$
\State $\mathcal{E}^{(\text{unmix})} = \{(u, v) \mid (u, v) \in \Delta \mathcal{E}, u, v \in \mathcal{V}^{(\text{unmix})}\}$
\State $G^{(\text{smp})} = (\mathcal{V}^{(\text{mix})} \cup \mathcal{V}^{(\text{unmix})}, \mathcal{E}^{(\text{mix})} \cup \mathcal{E}^{(\text{unmix})})$
\EndIf
\State $G' = G^{(\text{exp})} \cup G^{(\text{smp})}$ \Comment{Combine explanation and sampled subgraphs}
\State \textbf{return} $G'$
\end{algorithmic}
\end{flushleft}
\label{alg:mixup}
\end{algorithm}

\begin{algorithm}[h]
\caption{EPA-GRL Algorithm}
\begin{flushleft}
\textbf{Input:} Labeled dataset $\mathcal{T}_{\ell}$, unlabeled dataset $\mathcal{T}_u$, GNN pre-train epochs $e_w$, contrastive learning epochs $e_s$, temperature $\tau$, batch size $N$ \\
\textbf{Output:} Trained GNN encoder $f_\text{enc}(\cdot)$, trained explainer $\Psi(\cdot)$\\
\begin{algorithmic}[1]
\State Initialize GNN model $f\text{pt}(\cdot)$, encoder $f_\text{enc}(\cdot)$, and projection head $f_\text{pro}(\cdot)$
\For{epoch $= 1$ to $e_w$}
    \For{each $G_i \in \mathcal{T}_{\ell}$}
        \State Update $f_\text{pt}$ via the Cross-Entropy Loss,~Eq.(\ref{eq:ce})
    \EndFor
\EndFor
\State Train parametric GNN explainer $\Psi(\cdot)$ using $\mathcal{T}_{\ell}$ with Eq.(\ref{eq:explain})
\For{epoch $= 1$ to $e_s$}
    \For{sampled minibatch $\{G_i\}_{i=1}^N$ from $\mathcal{T}_{\ell} \cup \mathcal{T}_u$}
        \State Sample $t, t'$ from \{EPA-Node Dropping, EPA-Edge Dropping, EPA-Attribute Masking, EPA-Subgraph, EPA-Mixup\}
        \State $G_{i,1} = t(G_i, \Psi)$
        \State $G_{i,2} = t'(G_i, \Psi)$
    \EndFor
    \State Compute  self-supervised loss using Eq.(\ref{eq:graphcl}) or Eq.(\ref{eq:simsiamloss})
    \State Update $f_\text{enc}(\cdot)$ and $f_\text{pro}(\cdot)$ by minimizing self-supervised loss
\EndFor
\State \textbf{return} $f_\text{enc}(\cdot)$, $\Psi(\cdot)$
\end{algorithmic}
\end{flushleft}
\label{alg:graphrl}
\end{algorithm}

\section{C. Full Experimental Setups} \label{sec:fullexp}
\subsection{1. Datasets}
\textbf{{\mutag} \cite{luo2020parameterized}.}
{\mutag} dataset contains 4,337 molecules. These molecules are split into two groups based on their mutagenic effects on Salmonella Typhimurium, a Gram-negative bacterium. \\
\textbf{{\benz} \cite{agarwal2023evaluating}.}
{\benz} consists12,000 molecular graphs. They are divided into two categories: one where the molecules contain benzene rings and another where the benzene ring is absent. \\
\textbf{{\alk} \cite{agarwal2023evaluating}.}
The {\alk} has 4,326 molecular graphs. It is divided into two categories. The positive samples with ground-truth explanations include alkanes and carbonyl functional groups in the given molecules. \\
\textbf{{\fluo} \cite{agarwal2023evaluating}.} {\fluo} dataset contains 8,671 molecular graphs. The ground-truth explanation depends on the specific combination of fluoride atoms and carbonyl functional groups found in each molecule.\\
\textbf{{\dd} \cite{dobson2003distinguishing}.} It contains 1,178 protein graphs, classified into two binary categories: enzymes and non-enzymes. Each node in the graph represents an amino acid. If the distance between two amino acids is within 6 Angstroms, the corresponding nodes will connect to each other by edges.\\
\textbf{{\prot} \cite{dobson2003distinguishing, borgwardt2005protein}.} {\prot} dataset consists of 1,113 protein graphs, following the same construction method as {\dd}.\\
\textbf{{\bamo} \cite{luo2020parameterized}.} It contains 1,000 synthetic graphs generated from a standard Barabási-Albert (BA) model. The dataset is split into two groups: one set of graphs incorporates patterns resembling a house structure, while the other set integrates five-node cyclic motifs.  \\
\textbf{{\hiv} \cite{luong2024fragment}.} {\hiv} dataset consists of 41,127 molecular graphs, each representing a compound tested for its ability to inhibit HIV replication. The dataset is divided into two categories: one contains active molecules that can effectively inhibit HIV, and the other contains inactive molecules that lack inhibitory activity. \\
\textbf{{\RB} \cite{yanardag2015deep}.} The {\RB} consists of 2,000 graphs. Each graph represents a Reddit thread, where nodes denote users and edges indicate co-participation in the same discussion. \\

The statistics of all datasets are summarized in Table \ref{tab:dataset}.

\label{sec:extensive_data}
\begin{table*}[h]
\centering
\caption{Statistics of datasets used for graph classification task.}
\begin{tabular}{lcccccc}
\hline
Dataset  & Domain & \#Graphs & Avg.\#nodes & Avg.\#edges  & \#Feature & \#Classes \\
\hline
{\mutag}      &  Biochemical molecules      & 2,951            & 30.32           & 30.77      & 14 & 2              \\
{\benz}     & Biochemical molecules      & 12,000           &  20.58            & 43.65  & 14  & 2              \\
{\alk}     &  Biochemical molecules      & 4,326            & 21.13            & 44.95   & 14 & 2              \\
{\fluo}    &  Biochemical molecules     & 8,671            & 21.36            & 45.37                            & 14 & 2              \\
{\dd}      & Bioinformatics    & 1,178              & 284.32        & 715.66    & 89  & 2 \\
{\prot}    & Bioinformatics      & 1,113            & 39.06           & 72.82    & 3 & 2   \\
{\bamo}    & Synthetic   & 1,000            & 25              & 51               & 1             & 2              \\
{\hiv}    & Biochemical molecules      &  41127   &    25.51      &  54.93  & 9 &   2 \\
{\RB}    & Social networks      & 2000      &  429.63     &  497.75  & 13 &   2 \\
\hline
\end{tabular}
\label{tab:dataset}
\end{table*}

\subsection{2. Baselines}\label{app.baseline}

\textbf{{GraphCL} \cite{graphcl}.} GraphCL is a framework for unsupervised graph representation learning. It applies four types of graph augmentation methods: node dropping, edge dropping, attribute masking, and subgraph sampling.The code is available at \url{https://github.com/Shen-Lab/GraphCL}.\\
\textbf{{SimSiam} \cite{chen2021exploring}.} SimSiam does not require negative samples, large batches, or momentum encoders to train GNNs. It relies on a stop-gradient operation to prevent collapsing solutions. The code is available at \url{https://github.com/leaderj1001/SimSiam}.\\
\textbf{{AD-GCL} \cite{suresh2021adversarial}.} This framwork optimizes adversarial graph augmentations to prevent GNNs from learning redundant information. It uses trainable edge-dropping strategy to enhance the robustness of GNNs. The code is available at \url{https://github.com/susheels/adgcl}.\\
\textbf{{JOAO} \cite{you2021graph}.} JOAO uses min-max optimization to dynamically select data augmentation methods, allowing the augmentation strategy to adjust during training. The code is available at \url{https://github.com/Shen-Lab/GraphCL_Automated}.\\
\textbf{{AutoGCL} \cite{yin2022autogcl}.} It is an automatic graph contrastive learning framework that uses learnable generators with auto-augmentation to preserve key graph structures while introducing variance augmentation and jointly train the generator, encoder and classifier. The code is available at \url{https://github.com/Somedaywilldo/AutoGCL}.\\
\textbf{{SimGrace} \cite{xia2022simgrace}.} SimGrace does not require data augmentation. It uses the original graph as input and applies both the GNN and its perturbed version as two encoders to obtain two correlated views for contrastive learning. The code is available at \url{https://github.com/junxia97/SimGRACE}.\\
\textbf{GLA \cite{yue2022label}.} GLA uses graph contrastive learning to learn label-invariant features and improve supervision transfer between similar graphs. The code is available at \url{https://github.com/brandeis-machine-learning/Graph-Label-invariant-Augmentation}.\\
\textbf{ENGAGE \cite{shi2023engage}.} ENGAGE uses the Smoothed Activation Map (SAM) to identify key nodes from representation distributions and applies this to guide graph augmentation in contrastive learning. The code is available at \url{https://github.com/sycny/ENGAGE}.\\
\textbf{DRGCL \cite{ji2024rethinking}.} DRGCL incorporates dimension principle learning into graph contrastive learning. It uses a dedicated acquisition network to learn the dimension principle and applies a redundancy minimization loss to improve representation quality. The code is available at \url{https://github.com/ByronJi/DRGCL/tree/main}. \\
\textbf{CI-GCL \cite{tan2024community}.} It introduces a community-invariant contrastive learning framework that preserves the intrinsic community structure of graphs during augmentation. Community information is used to guide positive pair construction and maintain structural consistency. The code is available at \url{https://github.com/ShiyinTan/CI-GCL}. \\

\subsection{3. Implementation}\label{app.implementation}

A 3-layer Graph Convolutional Network (GCN) \cite{kipf2016semi} is used as the backbone GNN for all GRL methods. It is evaluated over 10 runs using random seeds from 0 to 9. For each seed, the dataset is randomly split into disjoint train/val/test sets in an 80\%/10\%/10\% ratio. We randomly sample 50 graphs from the training set for training SVM. For semi-supervised GRL methods (including \ours-GRL), the same 50 labeled graphs are used together with the unlabeled graphs in the training set for training GNNs. For unsupervised GRL methods, all graphs in the dataset (without using labels) are used for training GNNs. 

For all methods, the GNNs are trained using Adam optimizer \cite{kinga2015method} with a learning rate of $1 \times 10^{-3}$. For our \ours-GRL, a weight decay of $5 \times 10^{-4}$ is used for training the explainer, and the temperature parameter $\tau$ in Eq (\ref{eq:graphcl}) is set as 0.2. For the augmentation baselines, Node-Dropping removes 10\% nodes; Edge-Dropping perturbs 10\% edges; Attribute-Masking masks 10\% features; Subgraph-Sample randomly selects half of the nodes from the graph as starting nodes and performs random walks of 10 steps from each; Mix-up prunes 20\% of a graph randomly picked in the same batch to for an augmented graph. The implementation details about \ours\ augmentations are deferred to Appendix \ref{app.augalg}. All experiments are performed on a Linux machine with 8 Nvidia A100-PCIE 40GB GPUs, with CUDA version 12.4.

\section{D. Additional Experiments}

\subsection{1. Experiments with Another GRL Framework} \label{sec:extensive_framwork}

In Table \ref{tab:aucresults_simsiam}, we present results comparing \ours\ with various augmentation techniques under the SimSiam GRL framework. The results show that \ours\ consistently outperforms the Vanilla on most augmentation methods and datasets. For example, on {\fluo} and {\dd}, \ours\ achieves clear improvements, particularly with Node Dropping and Edge Dropping. These findings further demonstrate the generalizability and plug-and-play nature of \ours\ under different augmentation techniques.

\begin{table*}[h]
    \centering
    \caption{Comparison of different graph augmentation methods using SimSiam as the GRL framework.}
            \begin{tabular}{l|l|cccccc}
    \hline
    \multicolumn{2}{c|}{Augmentation Method} & {\mutag} & {\benz} & Alkane-Car. & Fluoride-Car. &{\dd} & {\prot}  \\ \hline
    \multirow{2}{*}{Node Dropping} & Vanilla  & 0.840$_{\pm 0.042}$ & 0.725$_{\pm 0.061}$ & 0.943$_{\pm 0.026}$  & 0.607$_{\pm 0.040}$ & 0.674$_{\pm 0.036}$ & 0.700$_{\pm 0.060}$ \\ 
                               & {\ours} & \textbf{0.855$_{\pm 0.049}$} & \textbf{0.730$_{\pm 0.066}$} & \textbf{0.945}$_{\pm 0.031}$  & \textbf{0.637$_{\pm 0.049}$} & \textbf{0.676}$_{\pm 0.037}$ & \textbf{0.731$_{\pm 0.090}$}\\     
    \hline
    \multirow{2}{*}{Edge Dropping } & Vanilla & 0.837$_{\pm 0.045}$ & 0.735$_{\pm 0.061}$ & 0.947$_{\pm 0.021}$  & 0.618$_{\pm 0.045}$ & \textbf{0.671}$_{\pm 0.056}$ & 0.687$_{\pm 0.089}$ \\
                    & {\ours} & \textbf{0.855$_{\pm 0.051}$} & \textbf{0.738$_{\pm 0.073}$} & \textbf{0.948}$_{\pm 0.030}$  & \textbf{0.640$_{\pm 0.043}$} & 0.668$_{\pm 0.031}$ & \textbf{0.728$_{\pm 0.076}$}  \\  
    \hline
    \multirow{2}{*}{Attribute Masking} & Vanilla & 0.840$_{\pm 0.048}$ & \textbf{0.715$_{\pm 0.052}$} & \textbf{0.947}$_{\pm 0.014}$ &  0.610$_{\pm 0.026}$ & 0.649$_{\pm 0.065}$ & 0.726$_{\pm 0.070}$ \\ 
                                   & {\ours}  &  \textbf{0.854$_{\pm 0.048}$} & 0.711$_{\pm 0.070}$ & 0.943$_{\pm 0.023}$  & \textbf{0.624$_{\pm 0.064}$} & \textbf{0.668}$_{\pm 0.059}$ & \textbf{0.750$_{\pm 0.080}$} \\ 
    \hline
    \multirow{2}{*}{Subgraph} & Vanilla & 0.838$_{\pm 0.039}$ & 0.704$_{\pm 0.079}$ & \textbf{0.950}$_{\pm 0.015}$  & 0.610$_{\pm 0.022}$ & 0.665$_{\pm 0.031}$ & 0.702$_{\pm 0.068}$ \\ 
                                   & {\ours} & \textbf{0.847$_{\pm 0.052}$} & \textbf{0.715$_{\pm 0.088}$} & 0.944$_{\pm 0.021}$ & \textbf{0.629$_{\pm 0.047}$} & \textbf{0.673}$_{\pm 0.074}$ & \textbf{0.744$_{\pm 0.084}$} \\ 
    \hline
    \multirow{2}{*}{Mixup} & Vanilla & 0.829$_{\pm 0.034}$ & \textbf{0.729$_{\pm 0.070}$} & 0.945$_{\pm 0.026}$  & 0.620$_{\pm 0.031}$ & 0.654$_{\pm 0.060}$ & 0.700$_{\pm 0.075}$ \\ 
                                   & {\ours}  & \textbf{0.830$_{\pm 0.070}$} & 0.721$_{\pm 0.062}$ & \textbf{0.951}$_{\pm 0.037}$  & \textbf{0.647$_{\pm 0.060}$} & \textbf{0.661}$_{\pm 0.060}$ & \textbf{0.721}$_{\pm 0.070}$ \\
    \hline
    \end{tabular}
    \label{tab:aucresults_simsiam}
\end{table*}

\subsection{2. Experiments with Another Augmentation Method}
\label{sec:extensive_aug}

To further demonstrate the generalizability of our method across different augmentation strategies, we incorporate the NodeSam augmentation proposed in \cite{yoo2022model} as an additional baseline. NodeSam modifies the graph structure by performing node split and merge operations in a balanced manner. We compare the standard NodeSam augmentation (Vanilla) with its counterpart integrated with our method (EPA) on all datasets. As shown in Table~\ref{tab:aucresults_nodesam_appendix}, EPA outperforms the Vanilla version, demonstrating its compatibility when applied to this augmentation method. 

\begin{table}[h]
    \centering
    \caption{Comparison of different graph augmentation methods with NodeSam Augmentation Method.}
        \begin{tabular}{l|l|cccccc}
    \hline
    \multicolumn{2}{c|}{Augmentation Method} & {\mutag} & {\benz} & Alkane-Car. & Fluoride-Car. &{\dd} & {\prot}  \\ \hline
    \multirow{2}{*}{NodeSam} & Vanilla  & 0.856$_{\pm 0.024}$ & 0.766$_{\pm 0.045}$ & 0.962$_{\pm 0.029}$ & 0.642$_{\pm 0.055}$ & 0.639$_{\pm 0.062}$ & 0.731$_{\pm 0.053}$ \\  
                                   & {\ours}     & \textbf{0.859}$_{\pm 0.017}$ & \textbf{0.770}$_{\pm 0.054}$ & \textbf{0.983}$_{\pm 0.020}$ & \textbf{0.669}$_{\pm 0.054}$ & \textbf{0.648}$_{\pm 0.053}$ & \textbf{0.734}$_{\pm 0.052}$\\     
    \hline
    \end{tabular}
    \label{tab:aucresults_nodesam_appendix}
\end{table}

\subsection{3. Experiments with Another Backbone}
\label{sec:extensive_backbone}

To evaluate the generalizability of our approach, we conduct additional experiments using a Graph Isomorphism Network (GIN)~\cite{xuhow2019} as the backbone encoder. We focus on the {\mutag} and {\prot} datasets for this analysis. Table \ref{tab:aucresults_gin_appendix} presents the results of these experiments.
We observe that {\ours} consistently outperforms the Vanilla augmentation methods in graph classification tasks on both datasets. This improvement suggests that the effectiveness of our explanation-preserving augmentation method is not limited to a specific graph neural network architecture. Instead, it appears to enhance the performance of different types of graph encoders, further validating the robustness and versatility of our approach.

\subsection{4. Experiments with Another Explainer Method}
\label{sec:extensive_exp}

To explore the impact of alternative explanation methods, we replace PGExplainer with Refine\cite{wang2021towards}. We conducted experiments on the {\mutag} and {\prot} datasets. The results are shown in Table~\ref{tab:aucresults_refine_appendix}. As shown, our method also improves the final classification performance when combined with Refine. This demonstrates that our method is compatible with various explanation methods. 

\begin{table}[h]
    \centering
    \caption{Comparison of different graph augmentation methods with GIN as the backbone.}
    \begin{tabular}{l|l|>{\centering\arraybackslash}m{2.5cm}>{\centering\arraybackslash}m{2.5cm}}
    \hline
    \multicolumn{2}{c|}{Augmentation Method} & {\mutag} & {\prot}  \\ \hline
    \multirow{2}{*}{Node Dropping} & Vanilla & 0.843$_{\pm 0.033}$ & 0.636$_{\pm 0.070}$ \\ 
                                   & {\ours}     & \textbf{0.846}$_{\pm 0.035}$ & \textbf{0.646}$_{\pm 0.074}$\\     
    \hline
    \multirow{2}{*}{Edge Dropping } & Vanilla & \textbf{0.836}$_{\pm 0.023}$ & 0.637$_{\pm 0.049}$ \\ 
                                   & {\ours}     & 0.831$_{\pm 0.033}$ & \textbf{0.650}$_{\pm 0.063}$  \\  
    \hline
    \multirow{2}{*}{Attribute Masking} & Vanilla & 0.827$_{\pm 0.040}$ & 0.622$_{\pm 0.057}$ \\ 
                                   & {\ours}     & \textbf{0.838}$_{\pm 0.027}$ & \textbf{0.656}$_{\pm 0.075}$ \\ 
    \hline
    \multirow{2}{*}{Subgraph} & Vanilla & 0.830$_{\pm 0.029}$ & 0.633$_{\pm 0.045}$ \\ 
                                   & {\ours}     & \textbf{0.839}$_{\pm 0.031}$ & \textbf{0.650}$_{\pm 0.051}$\\ 
    \hline
    \multirow{2}{*}{Mixup} & Vanilla & 0.842$_{\pm 0.038}$ & \textbf{0.625}$_{\pm 0.087}$ \\ 
                                   & {\ours}     & \textbf{0.846}$_{\pm 0.018}$ & 0.623$_{\pm 0.072}$ \\
    \hline
    \end{tabular}
    \label{tab:aucresults_gin_appendix}
\end{table}

\begin{table}[h]
    \centering
    \caption{Comparison of different graph augmentation methods with Refine as the explainer.}
    \begin{tabular}{l|l|>{\centering\arraybackslash}m{2.5cm}>{\centering\arraybackslash}m{2.5cm}}
    \hline
    \multicolumn{2}{c|}{Augmentation Method} & {\mutag} & {\prot}  \\ \hline
    \multirow{2}{*}{Node Dropping} & Vanilla & 0.803$_{\pm 0.030}$ & 0.728$_{\pm 0.073}$  \\ 
                                   & {\ours}     & \textbf{0.844}$_{\pm 0.039}$ & \textbf{0.730}$_{\pm 0.063}$ \\     
    \hline
    \multirow{2}{*}{Edge Dropping } & Vanilla & 0.858$_{\pm 0.027}$ & 0.702$_{\pm 0.077}$  \\ 
                                   & {\ours}     & \textbf{0.860}$_{\pm 0.029}$ & \textbf{0.728}$_{\pm 0.048}$ \\  
    \hline
    \multirow{2}{*}{Attribute Masking} & Vanilla & 0.820$_{\pm 0.064}$ & 0.683$_{\pm 0.077}$  \\ 
                                   & {\ours}   & \textbf{0.827}$_{\pm 0.046}$ & \textbf{0.713}$_{\pm 0.074}$ \\ 
    \hline
    \multirow{2}{*}{Subgraph} & Vanilla & 0.842$_{\pm 0.038}$ & 0.704$_{\pm 0.077}$  \\ 
                                   & {\ours}  & \textbf{0.844}$_{\pm 0.039}$ & \textbf{0.726}$_{\pm 0.044}$ \\ 
    \hline
    \multirow{2}{*}{Mixup} & Vanilla & 0.850$_{\pm 0.024}$ & 0.728$_{\pm 0.072}$  \\ 
                                   & {\ours}  & \textbf{0.853}$_{\pm 0.036}$ & \textbf{0.731}$_{\pm 0.067}$ \\
    \hline
    \end{tabular}
    \label{tab:aucresults_refine_appendix}
\end{table}

\subsection{5. Training Time Analysis}
\label{sec:timecost}

To evaluate the computational efficiency of our method, we report the total training time on several datasets in Table \ref{tab:time_cost}. We compare the training time of \ours\ with GraphCL and also account for the time required to train the explainer. Notably, the explanation-based augmentation step is independent of the GNN’s size or complexity, as the explainer is trained on a relatively small subset of graphs. As shown in the table, the explainer can be trained efficiently. Although \ours\ introduces an additional explanation phase, it results in only a modest increase in overall runtime compared to the baseline. This trade-off is acceptable, particularly considering the consistent performance improvements in classification performance achieved by our method.

\begin{table*}[h]
\centering
\caption{Training time of the explainer, vanilla (GraphCL), and our method (\ours).}
\begin{tabular}{lccc>{\centering\arraybackslash}m{2.2cm}}
\hline
  & Training Explainer (sec) & Vanilla (sec) & \ours\ (sec) & Runtime Ratio \\
\hline
{\mutag}    &  0.543   &  47      & 76            & 1.61×  \\
{\dd}      & 0.618 & 23    & 37              & 1.61×    \\
{\prot}    & 0.262  & 17      & 27            & 1.59×   \\
\hline
\end{tabular}
\label{tab:time_cost}
\end{table*}

\subsection{6. Comparison with GRL Methods}
In Table \ref{tab:baselines_2}, we compare \ours-GRL with the SOTA self-/semi-supervised GRL methods, including AD-GCL \cite{suresh2021adversarial}, JOAO \cite{you2021graph}, AutoGCL \cite{yin2022autogcl}, SimGRACE \cite{xia2022simgrace}, GLA \cite{yue2022label}, ENGAGE \cite{shi2023engage}, 
DRGCL \cite{ji2024rethinking}, and CI-GCL \cite{tan2024community}. As they are un-/self-supervised methods, we further extend them to semi-supervised settings using a two-step framework proposed in \cite{graphcl}: (1) pretrain a GNN encoder on all unlabeled graphs using each of the aforementioned self-supervised learning methods; (2) fine-tune the GNN encoder with the labeled graphs using cross-entropy loss. In particular, AutoGCL and CI-GCL has a design that enables joint training of the two processes. We adopt it for its semi-supervised training. 

All methods use a mixture of the aforementioned augmentation methods in the same way as in \cite{graphcl}, {\em i.e.}, randomly choosing an augmentation method for each original graph. All semi-supervised methods are trained using 50 labels. We also include ``Supervised'' as a baseline where the GNN encoder is only trained with the (small set of) labeled graphs $\gT_\ell$ using cross-entropy loss. We observe that using a few labeled graphs  $\gT_\ell$ in ``Supervised'' achieves competitive performances compared to training with the vast unlabeled graphs. 
Moreover, for each GRL baseline method, its semi-supervised version (``{\em SS-}'') is better than its self-supervised version in most cases. These results suggest the effectiveness of leveraging (a handful of) labels for GRL. Our method \ours-GRL integrates this merit but is remarkably different from the baseline methods which only use labels for updating model parameters. In contrast, \ours-GRL uses the limited amount of labels for structure learning, {\em i.e.}, explicitly learning the sub-structures that are semantically meaningful. Thus, it achieves substantial performance gains in general, meanwhile maintaining a consistently stable performance with low standard deviations.

\begin{table*}[t!]
\normalsize
\centering
\caption{Comparison of different self-supervised and semi-supervised (denoted as ``{\em SS-}'') GRL methods.}
\begin{tabular}{p{0.5cm}|p{2.5cm}|>{\centering\arraybackslash}m{2.5cm}>{\centering\arraybackslash}m{2.5cm}}\hline
\multicolumn{2}{c|}{GRL Method} & {\mutag} & {\prot} \\ \hline
\multicolumn{2}{c|}{Supervised} & 0.848$_{\pm 0.028}$ & 0.709$_{\pm 0.047}$ \\ \hline
\parbox[t]{2mm}{\multirow{8}{*}{\rotatebox[origin=c]{90}{Self-Sup.}}} & GraphCL & 0.846$_{\pm 0.032}$  & 0.712$_{\pm 0.050}$ \\
& AD-GCL & 0.854$_{\pm 0.025}$  & 0.736$_{\pm 0.070}$ \\
& JOAO & 0.823$_{\pm 0.034}$  & 0.725$_{\pm 0.067}$ \\
& AutoGCL & 0.812$_{\pm 0.071}$  & 0.722$_{\pm 0.053}$ \\
& SimGRACE & 0.834$_{\pm 0.050}$ & 0.719$_{\pm 0.068}$ \\ 
& ENGAGE & 0.840$_{\pm 0.044}$  & 0.700$_{\pm 0.063}$ \\ 
& DRGCL & 0.841$_{\pm 0.029}$  & 0.738$_{\pm 0.059}$ \\ 
& CI-GCL & 0.848$_{\pm 0.035}$  & 0.739$_{\pm 0.061}$ \\ \hline
\parbox[t]{2mm}{\multirow{9}{*}{\rotatebox[origin=c]{90}{Semi-Sup.}}} & {\em SS}-GraphCL & 0.853$_{\pm 0.034}$ & 0.720$_{\pm 0.063}$ \\
& {\em SS}-AD-GCL & 0.857$_{\pm 0.036}$ & 0.733$_{\pm 0.061}$ \\
& {\em SS}-JOAO & 0.849$_{\pm 0.038}$ & 0.726$_{\pm 0.063}$ \\
& {\em SS}-AutoGCL & 0.824$_{\pm 0.038}$ & 0.734$_{\pm 0.066}$ \\
& {\em SS}-SimGRACE & 0.834$_{\pm 0.041}$ & 0.713$_{\pm 0.056}$ \\ 
& GLA & \underline{0.859}$_{\pm 0.034}$ & 0.720$_{\pm 0.063}$ \\  
& {\em SS}-DRGCL & 0.846$_{\pm 0.031}$ & 0.735$_{\pm 0.064}$ \\ 
& {\em SS}-CI-GCL & 0.845$_{\pm 0.041}$ & \underline{0.740}$_{\pm 0.030}$ \\ \cline{2-4}
& {\ours}-GRL & \textbf{0.861}$_{\pm 0.032}$ & \textbf{0.744}$_{\pm 0.065}$\\ \hline
\end{tabular}
\label{tab:baselines_2}
\end{table*}

\subsection{7. Additional Ablation Studies} \label{app.ablation}

\begin{figure*}[h]
    \centering
    \begin{subfigure}[b]{0.24\textwidth}
        \centering
        \includegraphics[width=0.95\textwidth]{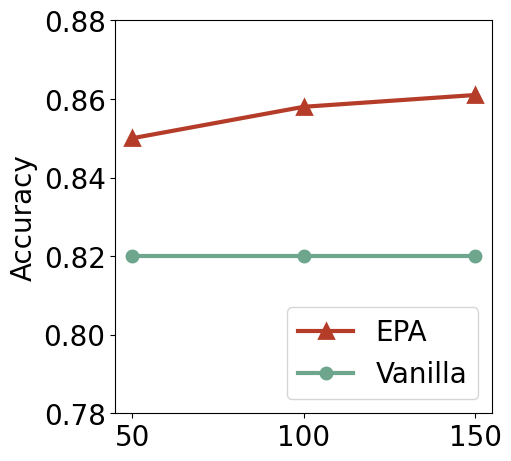}
        \caption{Attribute Masking}
        \label{pre_featuremask}
    \end{subfigure}
    \hspace{0.8cm}
    \begin{subfigure}[b]{0.24\textwidth}
        \centering
        \includegraphics[width=0.95\textwidth]{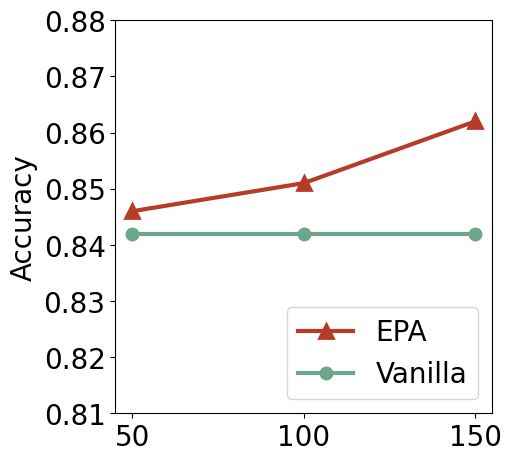}
        \caption{Subgraph}
        \label{pre_subgraph}
    \end{subfigure}
    \hspace{0.8cm}
    \begin{subfigure}[b]{0.24\textwidth}
        \centering
        \includegraphics[width=0.95\textwidth]{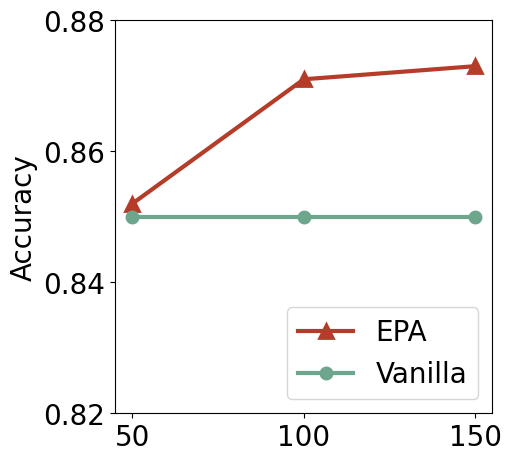}
        \caption{Mixup}
        \label{pre_mixup}
    \end{subfigure}
    \caption{Parameter analysis with different numbers of training samples for pre-training.}
    \label{fig:parameters_2_all}
\end{figure*}

\begin{figure*}[h]
    \centering
    \begin{subfigure}[b]{0.24\textwidth}
        \centering
        \includegraphics[width=0.95\textwidth]{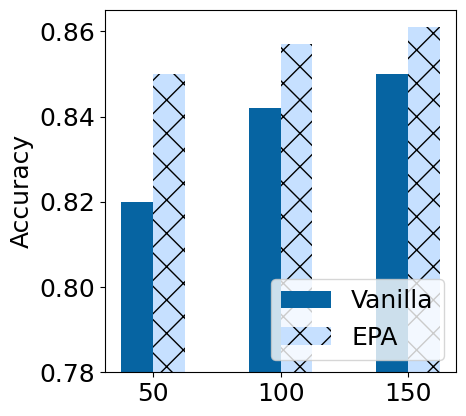}
        \caption{Attribute Masking}
        \label{featuremask}
    \end{subfigure}
    \hspace{0.8cm}
    \begin{subfigure}[b]{0.24\textwidth}
        \centering
        \includegraphics[width=0.95\textwidth]{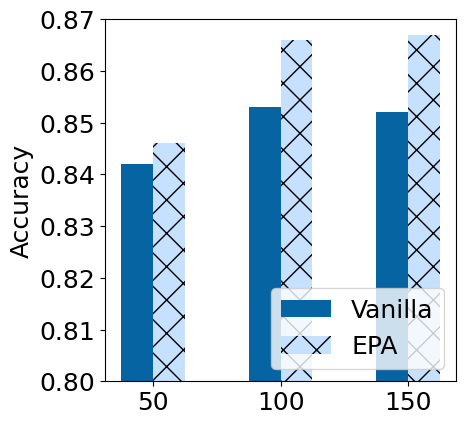}
        \caption{Subgraph}
        \label{subgraph}
    \end{subfigure}
    \hspace{0.8cm}
    \begin{subfigure}[b]{0.24\textwidth}
        \centering
        \includegraphics[width=0.95\textwidth]{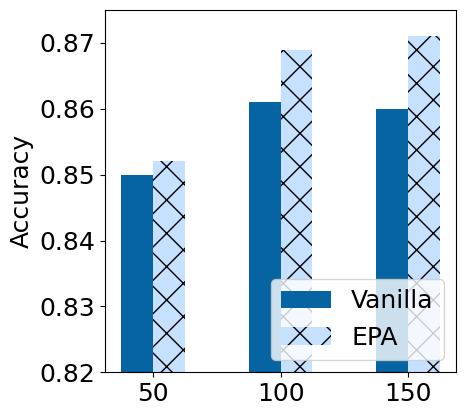}
        \caption{Mixup}
        \label{mixup}
    \end{subfigure}
    \caption{Parameter analysis with different number of training samples for downstream training.}
    \label{fig:parameters_1_all}
\end{figure*}

In Fig. \ref{fig:parameters_2_all} and Fig. \ref{fig:parameters_1_all}, we conduct a comprehensive ablation study on {\mutag} to evaluate the impact of the number of labeled graphs used for explainer pre-training and the downstream training of SVM on the Attribute Masking, Subgraph, and Mixup augmentation methods, respectively. The results indicate that {\ours} can use additional labels to generate improved augmentations, and \ours-GRL consistently outperforms the baseline in producing higher-quality embeddings.

\subsection{8. Robust Fidelity Metrics for Explanation Quality} 
\label{app.noise}
Traditional fidelity metrics for graph explanations rely on comparing model predictions between the original graph and various subgraphs \cite{liu2021dig, amara2022graphframex}. The standard positive fidelity metric $Fid_{+}$ measures prediction changes when removing the explanatory subgraph, while negative fidelity $Fid_{-}$ examines predictions on the explanatory subgraph alone \cite{pope2019explainability, yuan2022explainability}. However, these approaches often fail in practice because removing edges can create Out Of Distribution (OOD) samples, leading to unreliable predictions.

A recent work \cite{luo2024towards} addresses this limitation by introducing distribution-robust fidelity metrics $Fid_{\alpha_1,+}$ and $Fid_{\alpha_2,-}$. Given an input graph $G$ with label $y$, a classifier $f(\cdot)$, and an explanation subgraph $G^{(\text{exp})}$, these metrics are defined as:
\begin{align*}
   &\label{eq:fid_np} Fid_{\alpha_1,+} \triangleq f(G)_y - \mathbb{E}f(G-E_{\alpha_1}(G^{(\text{exp})}))_y,\\
   & Fid_{\alpha_2,-}\triangleq  f(G)_y - \mathbb{E}f(G^{(\text{exp})}+E_{\alpha_2}(G-G^{(\text{exp})}))_y,
\end{align*}
where $E_{\alpha}:G\mapsto G_{\alpha}$ is a stochastic graph sampling function with edge erasure probability $\alpha\in [0,1]$. It takes a graph $G$ as input and outputs a sampled graph $G_{\alpha}$ that retrains the same node set as $G$, with each edge independently included with probability $\alpha$. These metrics are designed to be robust against OOD issues in a wide range of scenarios. Therefore, we adopt these improved fidelity metrics as evaluation criteria in our experiments, using default settings $\alpha_1=0.1, \alpha_2=0.9$.

\begin{figure}[h]
    \centering
    \includegraphics[width=0.45\textwidth]{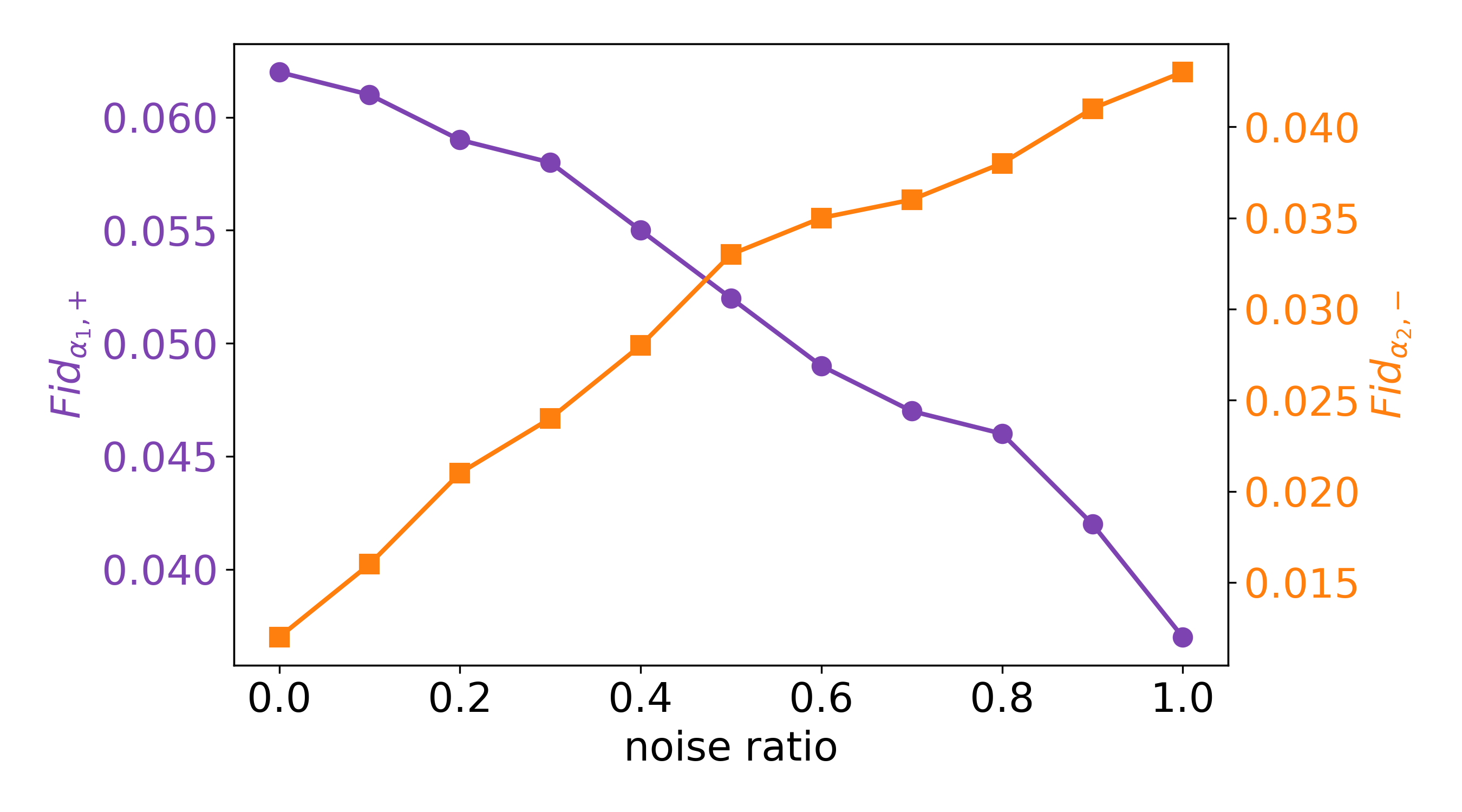}
    \caption{Fidelity values under different noise levels on {\mutag}.}
    \label{fig:noise_fid}
    \end{figure}

Fig. \ref{fig:noise_fid} demonstrates the effectiveness of these robust fidelity metrics in our experimental setup. When we systematically introduce noise into explanation subgraphs, $Fid_{\alpha_1,+}$ and $Fid_{\alpha_2,-}$ respond as theoretically expected: $Fid_{\alpha_1,+}$ decreases as the explanatory power of the noisy subgraphs diminishes, while $Fid_{\alpha_2,-}$ increases as these subgraphs become less sufficient for prediction. This consistent behavior under controlled noise validates the reliability of these metrics for evaluating explanation quality in our framework, supporting their use as quantitative measures in our main experiments. Using these metrics, we can verify that high-quality explanations (those with high $Fid_{\alpha_1,+}$ and low $Fid_{\alpha_2,-}$ indeed preserve the essential graph semantics. When such explanations guide our augmentation process, they lead to more effective contrastive learning, as evidenced by the improved classification performance shown in Fig. \ref{fig:correlation}. This relationship between explanation fidelity and model performance provides quantitative support for our semantics-preserving approach to graph augmentation.

\subsection{9. Hyperparameter Analysis} \label{app.hyperparameter}

\begin{figure}[h!]
    \centering
    \includegraphics[width=0.4\textwidth]{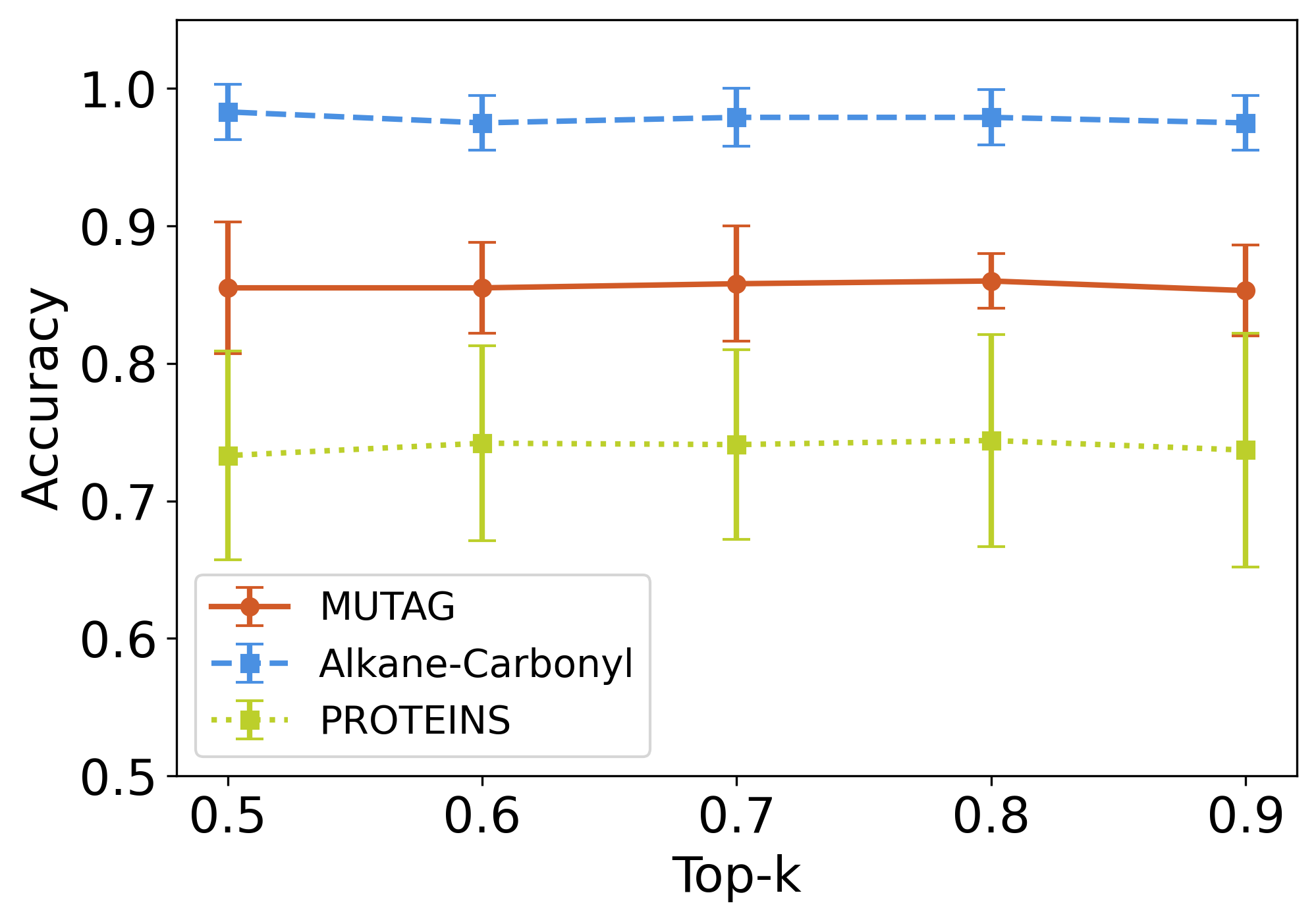}
    \caption{Hyperparameter analysis of k on {\mutag}, {\alk}, and {\prot}.} 
    \label{fig:hyper}
\end{figure}

In this section, we investigate the hyperparameter k which controls the size of the explanation subgraph using the {\mutag}, {\alk} and {\prot} datasets. As shown in Fig. \ref{fig:hyper}, the accuracy for {\mutag} remains stable for all values of k. For {\alk}, the accuracy improves slightly as k increases to 0.7 and then declines marginally. For {\prot}, the accuracy remains relatively stable, with minor fluctuations across different subgraph sizes. Despite these variations, the overall performance does not fluctuate significantly. In this study, we set k $=0.8$ for all datasets, as it ensures a balance between stable performance and robustness under different conditions.

\subsection{10. Case Studies} \label{app.case_study}

\begin{figure}[h]
    \captionsetup{type=table}
    \centering
    \caption{Visualization of the augmented graphs using different methods. Yellow edges highlight the explanatory sub-structures. Gray nodes denote deleted nodes. Dashed edges denote deleted edges.}
    \includegraphics[width=0.65\textwidth]{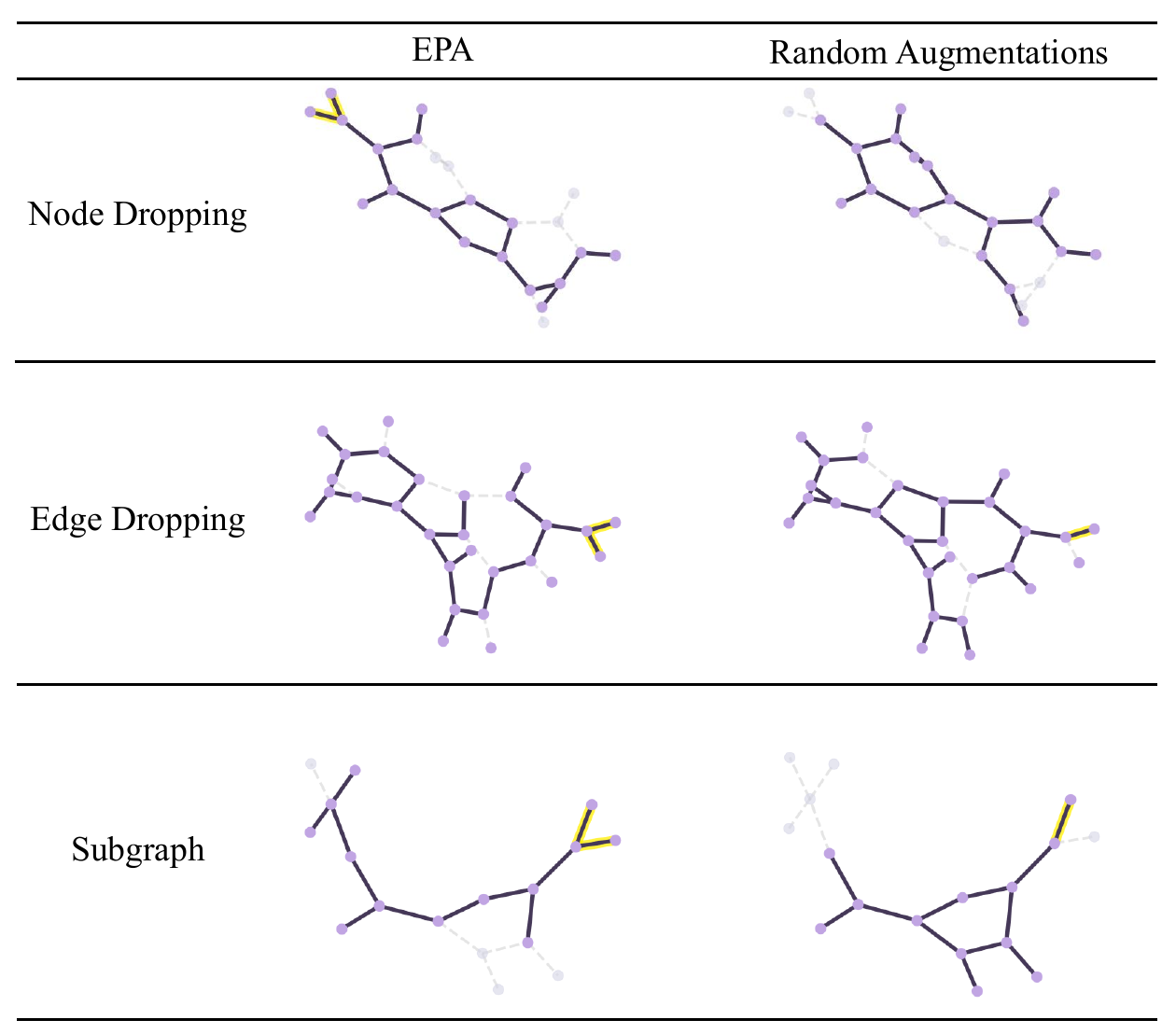}
    \label{table:casestudy}
\end{figure}

In this section, we investigate how \ours\ preserves semantics by examining the augmented graphs. Using {\mutag}, we visualize the augmented graphs for Node-Dropping, Edge-Dropping, and Subgraph Sampling as examples in Table \ref{table:casestudy}. The ground-truth explanatory sub-structures, {\em i.e.}, semantic patterns, are highlighted by yellow edges. As can be seen, through the GNN explainer, \ours\ successfully preserves the explanatory sub-structures meanwhile perturbing the marginal subgraphs (as represented by gray nodes and dashed edges). In contrast, Vanilla augmentations may randomly perturb the explanatory sub-structures, leading to a significant loss of semantics in the augmented graphs. In particular, it drops the entire semantic subgraph by Node Dropping. This result justifies the superior performance of \ours-GRL in Table \ref{tab:aucresults_graphcl} and Table \ref{tab:aucresults_simsiam}.

\subsection{11. Visualization of Graph Representations} \label{sec:vis}
\begin{figure}[h]
    \centering
    \begin{subfigure}[b]{0.28\textwidth}
        \centering
        \includegraphics[width=1.0\textwidth]{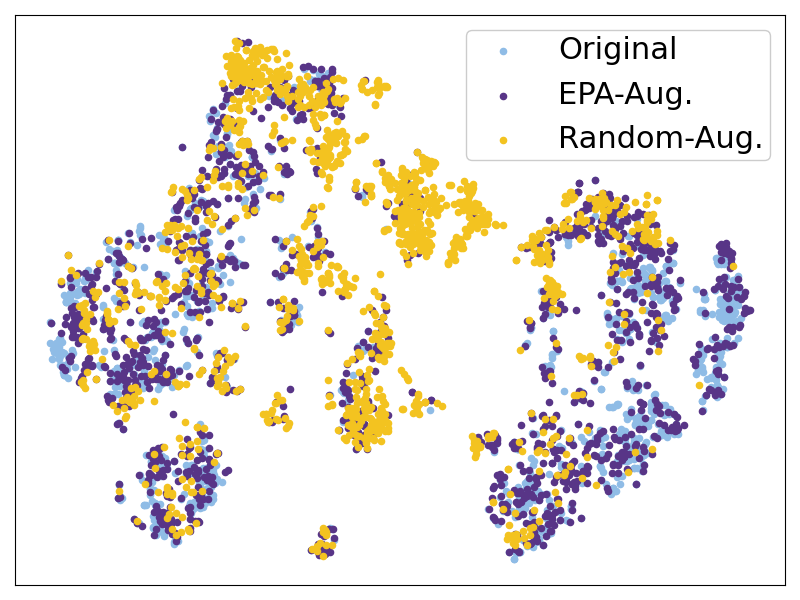}
        \caption{{\mutag}}
    \end{subfigure}
    \begin{subfigure}[b]{0.28\textwidth}
        \centering
        \includegraphics[width=1.0\textwidth]{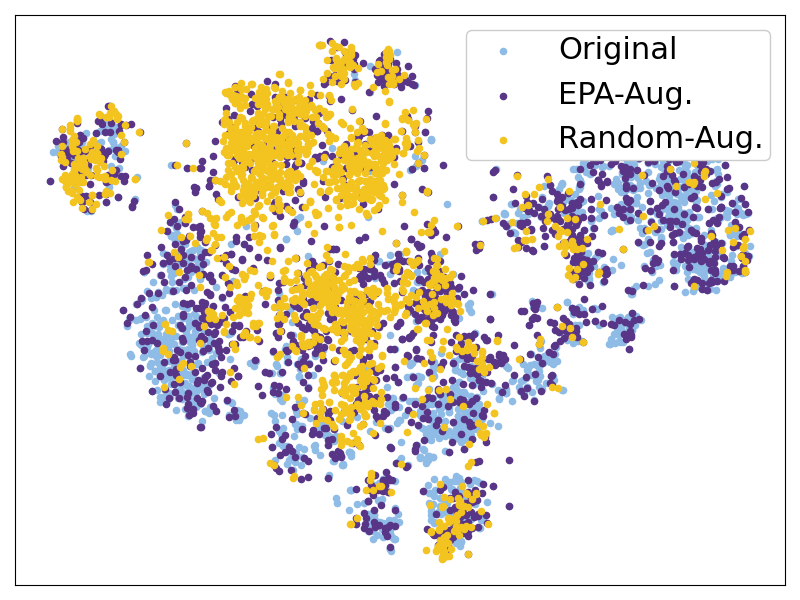}
        \caption{{\benz}}
    \end{subfigure}
    \begin{subfigure}[b]{0.28\textwidth}
        \centering
        \includegraphics[width=1.0\textwidth]{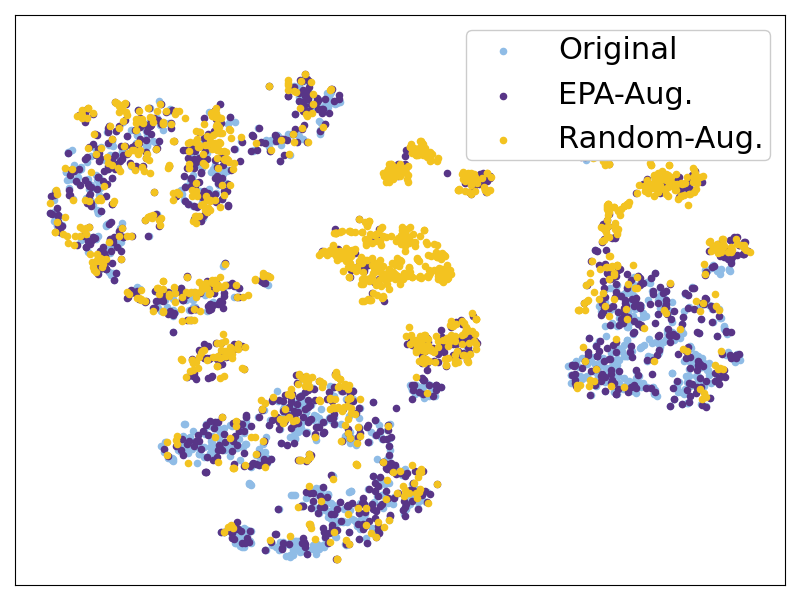}
        \caption{{\alk}}
    \end{subfigure}
    \begin{subfigure}[b]{0.28\textwidth}
        \centering
        \includegraphics[width=1.0\textwidth]{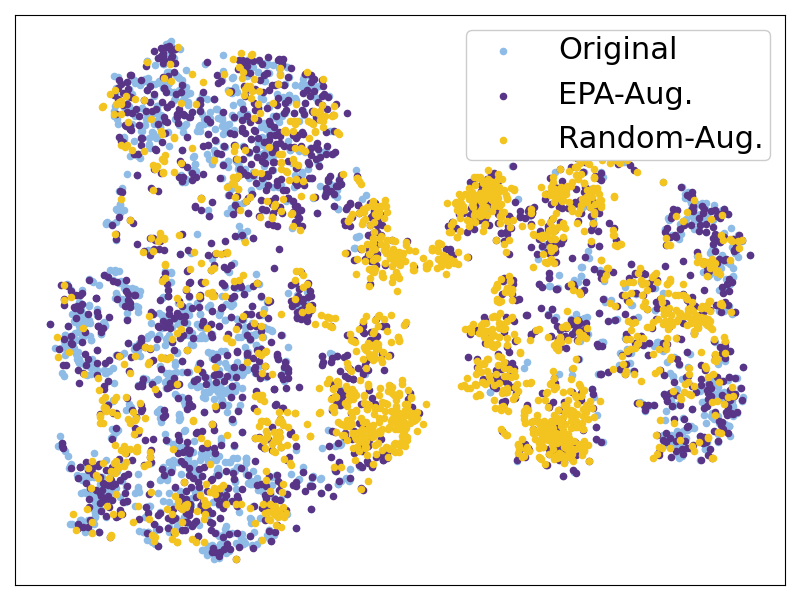}
        \caption{{\fluo}}
    \end{subfigure}
    \begin{subfigure}[b]{0.28\textwidth}
        \centering
        \includegraphics[width=1.0\textwidth]{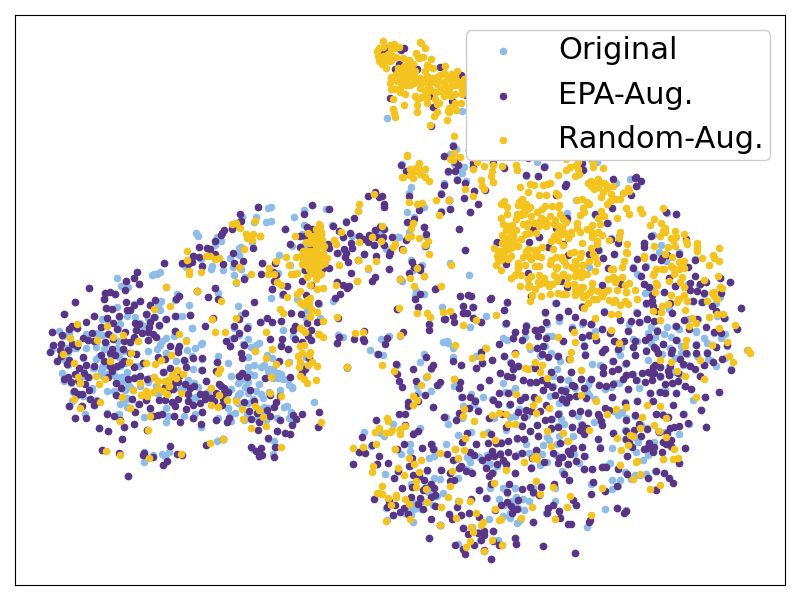}
        \caption{{\dd}}
    \end{subfigure}
    \begin{subfigure}[b]{0.28\textwidth}
        \centering
        \includegraphics[width=1.0\textwidth]{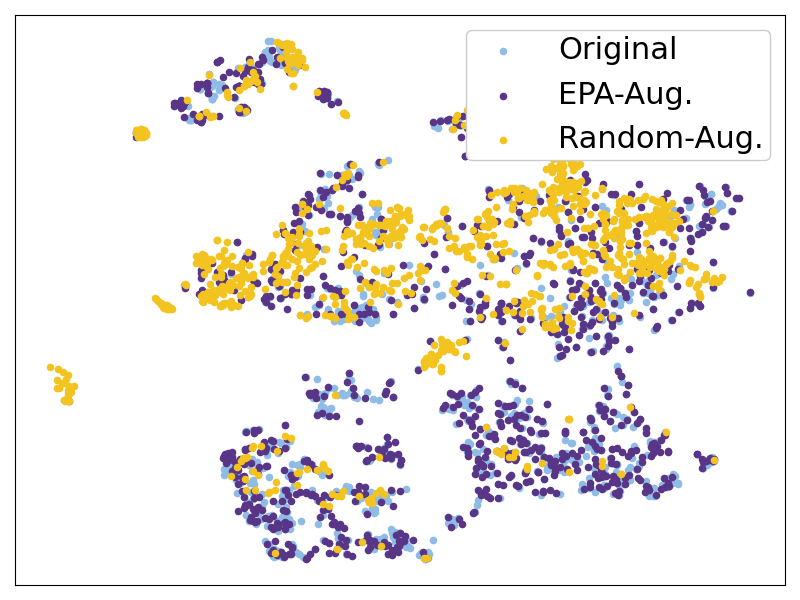}
        \caption{{\prot}}
    \end{subfigure}
    \caption{Visualizations of graph representations on all datasets.}
    \label{fig:full-vis}
\end{figure}

We present visual examples from all real datasets: {\mutag}, {\benz}, {\alk}, {\fluo}, {\dd}, and {\prot} to demonstrate the semantic preservation ability of different augmentation techniques. The visualizations are shown in Fig. \ref{fig:full-vis}. From the figure, we observe that the distribution difference between the original graph and the ``Random-Aug'' graph is large, while the difference between the original graph and the ``EPA-Aug'' graph is small. Although ``Random-Aug'' provides diverse augmentations, it may change key properties of the graph, affecting downstream tasks that rely on consistent graph semantics. This highlights the importance of semantic-preserving augmentations.

\end{document}